\documentclass[journal]{IEEEtran}
\usepackage{algorithmic}
\usepackage{amsthm}
\usepackage{amsmath}
\usepackage{amssymb}
\usepackage{array}
\usepackage[english]{babel}
\usepackage{bibentry}
\usepackage{cite}
\usepackage{color}
\usepackage{rotating}
\usepackage{graphicx}
\usepackage{enumitem}
\usepackage[T1]{fontenc}
\usepackage{url}
\usepackage{fancyref}
\usepackage{hyperref}
\usepackage{tabularx}
\usepackage{xspace}
\usepackage{subcaption}
\usepackage{breakurl}
\usepackage{multirow}

\newcommand{\eg}{\textit{e}.\textit{g}.\@\xspace}
\newcommand{\ie}{\textit{i}.\textit{e}.\@\xspace}
\newcommand{\etal}{\textit{et al}.\@\xspace}

\newcolumntype{Y}{>{\centering\arraybackslash}X}
\usepackage[font=footnotesize,labelfont=bf]{caption}

\begin{document}
\title{Adversarial Examples: Attacks and Defenses for Deep Learning\thanks{This work is partially supported by National Science Foundation (grants CNS-1842407, CNS-1747783, CNS-1624782, and OAC-1229576).}}

\author{Xiaoyong Yuan, Pan He, Qile Zhu, Xiaolin Li\IEEEauthorrefmark{1}\thanks{\IEEEauthorrefmark{1} Corresponding author.}
\\National Science Foundation Center for Big Learning, University of Florida
\\ \{chbrian, pan.he, valder\}@ufl.edu, andyli@ece.ufl.edu}
\maketitle

\IEEEpeerreviewmaketitle

\begin{abstract}
With rapid progress and significant successes in a wide spectrum of applications, deep learning is being applied in many safety-critical environments. However, deep neural networks have been recently found vulnerable to well-designed input samples, called \textit{adversarial examples}. Adversarial examples are imperceptible to human but can easily fool deep neural networks in the testing/deploying stage. The vulnerability to adversarial examples becomes one of the major risks for applying deep neural networks in safety-critical environments. Therefore, attacks and defenses on adversarial examples draw great attention. 
In this paper, we review recent findings on adversarial examples for deep neural networks, summarize the methods for generating adversarial examples, and propose a taxonomy of these methods. Under the taxonomy, applications for adversarial examples are investigated. We further elaborate on countermeasures for adversarial examples and explore the challenges and the potential solutions. 
\end{abstract}

\begin{IEEEkeywords}
deep neural network, deep learning, security, adversarial examples
\end{IEEEkeywords}

\section{Introduction}
\label{sec:intro}
Deep learning (DL) has made significant progress in a wide domain of machine learning (ML): image classification, object recognition~\cite{krizhevsky2012imagenet, simonyan2014very}, object detection~\cite{redmon2016yolo9000, ren2015faster}, speech recognition~\cite{saon2015ibm}, language translation~\cite{sutskever2014sequence}, voice synthesis~\cite{oord2016wavenet}. 
The online Go Master (AlphaGo~\cite{silver2016mastering}) beat more than 50 top go players in the world. Recently AlphaGo Zero~\cite{silver2017mastering} surpassed its previous version without using human knowledge and a generic version, AlphaZero~\cite{silver2017masteringchess}, achieved a superhuman level within 24 hours of cross domains of chess, Shogi, and Go.

Driven by the emergence of big data and hardware acceleration, deep learning requires less hand engineered features and expert knowledge. The intricacy of data can be extracted with higher and more abstract level representation from raw input features~\cite{lecun2015deep}. 

Constantly increasing number of real-world applications and systems have been powered by deep learning. For instance, companies from IT to the auto industry (\eg, Google, Telsa, Mercedes, and Uber) are testing self-driving cars, which require plenty of deep learning techniques such as object recognition, reinforcement learning, and multimodal learning. Face recognition system has been deployed in ATMs as a method of biometric authentication~\cite{middlehurst2015}. Apple also provides face authentication to unlock mobile phones~\cite{faceid}. Behavior-based malware detection and anomaly detection solutions are built upon deep learning to find semantic features~\cite{dahl2013large,yuan2014droid,saxe2015deep,sun2017learning}. 

Despite great successes in numerous applications, many of deep learning empowered applications are life crucial, raising great concerns in the field of safety and security. ``With great power comes great responsibility''~\cite{spider}. Recent studies find that deep learning is vulnerable against well-designed input samples. These samples can easily fool a well-performed deep learning model with little perturbations imperceptible to humans.

Szegedy \etal first generated small perturbations on the images for the image classification problem and fooled state-of-the-art deep neural networks with high probability~\cite{szegedy2013intriguing}. These misclassified samples were named as \textit{Adversarial Examples}. 

Extensive deep learning based applications have been used or planned to be deployed in the physical world, especially in the safety-critical environments. In the meanwhile, recent studies show that adversarial examples can be applied to real world. For instance, an adversary can construct physical adversarial examples and confuse autonomous vehicles by manipulating the stop sign in a traffic sign recognition system~\cite{kurakin2016adversarial,evtimov2017robust} or removing the segmentation of pedestrians in an object recognition system~\cite{xie2017adversarial}. Attackers can generate adversarial commands against automatic speech recognition (ASR) models and Voice Controllable System (VCS)~\cite{carlini2016hidden,zhang2017dolphinatack} such as Apple Siri~\cite{siri}, Amazon Alexa~\cite{alexa}, and Microsoft Cortana~\cite{cortana}. 

Deep learning is widely regarded as a ``black box'' technique --- we all know that it performs well, but with limited knowledge of the reason~\cite{knight2017dark,castelvecchi2016can}. Many studies have been proposed to explain and interpret deep neural networks~\cite{koh2017understanding,lipton2016mythos,shwartz2017opening,selvaraju2016grad}. From inspecting adversarial examples, we may gain insights on semantic inner levels of neural networks~\cite{lu2017safetynet} and find problematic decision boundaries, which in turn helps to increase robustness and performance of neural networks~\cite{wu2017adversarial} and improve the interpretability~\cite{dong2017towards}.


In this paper, we investigate and summarize  approaches for generating adversarial examples, applications for adversarial examples and corresponding countermeasures. We explore the characteristics and  possible causes of adversarial examples. Recent advances in deep learning revolve around supervised learning, especially in the field of computer vision task. Therefore, most adversarial examples are generated against computer vision models. We mainly discuss  adversarial examples for image classification/object recognition tasks in this paper. Adversarial examples for other tasks will be investigated in Section~\ref{sec:app}. 

Inspired by~\cite{papernot2016towards}, we define the \textbf{Threat Model} as follows:
\begin{itemize}
        \item  The adversaries can attack only at the testing/deploying stage. They can tamper only the input data in the testing stage after the victim deep learning model is trained. Neither the trained model or the training dataset can be modified. The adversaries may have knowledge of trained models (architectures and parameters) but are not allowed to modify models, which is a common assumption for many online machine learning services. Attacking at the training stage (\eg, training data poisoning) is another interesting topic and has been studied in~\cite{biggio2012poisoning,roli2013pattern,xiao2015feature,mozaffari2015systematic,beatson2016blind,alfeld2016data}. Due to the limitation of space, we do not include this topic in the paper. 
    \item  We focus on attacks against models built with deep neural networks, due to their great performance achieved. We will discuss adversarial examples against conventional machine learning (\eg, SVM, Random Forest) in Section~\ref{sec:back}. Adversarial examples against deep neural networks proved effective in conventional machine learning models~\cite{papernot2016transferability} (see Section~\ref{sec:transfer}). 
    \item Adversaries only aim at compromising \textit{integrity}. Integrity is presented by performance metrics (\eg, accuracy, F1 score, AUC), which is essential to a deep learning model. Although other security issues pertaining to confidentiality and privacy have been drawn attention in deep learning~\cite{fredrikson2015model,abadi2016deep,shokri2017membership}, we focus on the attacks that degrade the performance of deep learning models, cause an increase of false positives and false negatives.
    \item The rest of the threat model differs in different adversarial attacks. We will categorize them in Section~\ref{sec:taxonomy}.
\end{itemize}

Notations and symbols used in this paper are listed in Table~\ref{tab:notation}.
\begin{table}[!bt]
\caption{Notation and symbols used in this paper}
\label{tab:notation}
\begin{tabularx}{\linewidth}{|c|X|}
    \hline
    Notations and Symbols & Description \\ \hline
    $x$     & original (clean, unmodified) input data \\
    $l$     & label of class in the classification problem. $l=1,2, \ldots, m$, where $m$ is the number of classes\\
    $x'$    & adversarial example (modified input data) \\
    $l'$    & label of the adversarial class in targeted adversarial examples \\
    $f(\cdot)$     & deep learning model (for the image classification task, $f \in F: \mathbb{R}^n \rightarrow l$) \\
    $\theta$     & parameters of deep learning model $f$\\
    $J_f(\cdot, \cdot)$     & loss function (\eg, cross-entropy) of model $f$ \\ 
    $\eta$     & difference between original and modified input data: $\eta = x'-x$ (the exact same size as the input data)\\
    $\|\cdot\|_p$   & $\ell_p$ norm\\
    $\nabla$    & gradient \\
    $H(\cdot)$  & Hessian, the second-order of derivatives \\
    $KL(\cdot)$    & Kullback-Leibler (KL) divergence function\\\hline
\end{tabularx}
\end{table}





This paper presents the following contributions:
\begin{itemize}
    \item To systematically analyze approaches for generating adversarial examples, we taxonomize attack approaches along different axes to provide an accessible and intuitive overview of these approaches.
    \item We investigate recent approaches and their variants for generating adversarial examples and compare them using the proposed taxonomy. We show examples of selected applications from fields of reinforcement learning, generative modeling, face recognition, object detection, semantic segmentation, natural language processing, and malware detection.  Countermeasures for adversarial examples are also discussed.
    \item We outline main challenges and potential future research directions for adversarial examples based on three main problems: transferability of adversarial examples, existence of adversarial examples, and robustness evaluation of deep neural networks.
\end{itemize}


The remaining of this paper is organized as follows. Section~\ref{sec:back} introduces the background of deep learning techniques, models, and datasets. We discuss adversarial examples raised in conventional machine learning in Section~\ref{sec:back}. We propose a taxonomy of approaches for generating adversarial examples in Section~\ref{sec:taxonomy} and elaborate on these approaches in Section~\ref{sec:method}. In Section~\ref{sec:app}, we discuss applications for adversarial examples. Corresponding countermeasures are investigated in Section~\ref{sec:counter}. We discuss current challenges and potential solutions in Section~\ref{sec:analysis}. Section~\ref{sec:conclusion} concludes the work.

\section{Background}
\label{sec:back}
In this section, we briefly introduce basic deep learning techniques and approaches related to adversarial examples. Next, we review adversarial examples in the era of conventional ML and compare the difference between adversarial examples in conventional ML and that in DL.

\subsection{Brief Introduction to Deep Learning}

This subsection discusses main concepts, existed techniques, popular architectures, and standard datasets in deep learning, which, due to the extensive use and breakthrough successes, have become acknowledged targets of attacks, where adversaries are usually applied to evaluate their attack methods.

\subsubsection{Main concepts in deep learning}

Deep learning is a type of machine learning methods that makes computers to learn from experience and knowledge without explicit programming and extract useful patterns from raw data. For conventional machine learning algorithms, it is difficult to extract well-represented features due to limitations, such as curse of dimensionality~\cite{bengio2007scaling}, computational bottleneck~\cite{storcheus2015survey}, and requirement of the domain and expert knowledge. Deep learning solves the problem of representation by building multiple simple features to represent a sophisticated concept. For example, a deep learning-based image classification system represents an object by describing edges, fabrics, and structures in the hidden layers. With the increasing number of available training data, deep learning becomes more powerful. Deep learning models have solved many complicated problems, with the help of hardware acceleration in computational time. 

A neural network layer is composed of a set of perceptrons (artificial neurons). Each perceptron maps a set of inputs to output values with an activation function. The function of a neural network is formed in a chain:
\begin{equation}
f(x) = f^{(k)}(\cdots f^{(2)}(f^{(1)}(x))),
\end{equation}
where $f^{(i)}$ is the function of the $i$th layer of the network, $i=1,2,\cdots k$. 

Convolutional neural networks (CNNs) and Recurrent neural networks (RNNs) are two most widely used neural networks in recent neural network architectures. CNNs deploy convolution operations on hidden layers to share weights and reduce the number of parameters. CNNs can extract local information from grid-like input data. CNNs have shown incredible successes in computer vision tasks, such as image classification~\cite{krizhevsky2012imagenet,he2015delving}, object detection~\cite{redmon2016you, ren2015faster}, and semantic segmentation~\cite{long2015fully,chen2017rethinking}. RNNs are neural networks for processing sequential input data with variable length. RNNs produce outputs at each time step. The hidden neuron at each time step is calculated based on current input data and hidden neurons at previous time step. Long Short-Term Memory (LSTM) and Gated Recurrent Unit (GRU) with controllable gates are designed to avoid vanishing/exploding gradients of RNNs in long-term dependency.

\subsubsection{Architectures of deep neural networks}
Several deep learning architectures are widely used in computer vision tasks: \textit{LeNet}~\cite{le2013building}, \textit{VGG}~\cite{simonyan2014very}, \textit{AlexNet}~\cite{krizhevsky2012imagenet}, \textit{GoogLeNet}~\cite{szegedy2015going,szegedy2016rethinking,szegedy2017inception} (Inception V1-V4), and \textit{ResNet}\cite{he2015delving}, from the simplest (oldest) network to the deepest and the most complex (newest) one. \textit{AlexNet} first showed that deep learning models can largely surpass conventional machine learning algorithms in the ImageNet 2012 challenge and led the future study of deep learning. These architectures made tremendous breakthroughs in the ImageNet challenge and can be seen as milestones in image classification problem. Attackers usually generate adversarial examples against these baseline architectures. 

\subsubsection{Standard deep learning datasets}
\textit{MNIST}, \textit{CIFAR-10}, \textit{ImageNet} are three widely used datasets in computer vision tasks. The \textit{MNIST} dataset is for handwritten digits recognition~\cite{lecun1998mnist}. The \textit{CIFAR-10} dataset and the \textit{ImageNet} dataset are for image recognition task~\cite{krizhevsky2009learning}. The \textit{CIFAR-10} consists of 60,000 tiny color images ($32\times32$) with ten classes. The \textit{ImageNet} dataset consists 14,197,122 images with 1,000 classes~\cite{russakovsky2015ImageNet}. Because of the large number of images in the \textit{ImageNet} dataset, most adversarial approaches are evaluated on only part of the \textit{ImageNet} dataset. The \textit{Street View House Numbers (SVHN)} dataset, similar to the \textit{MNIST} dataset, consists of ten digits obtained from real-world house numbers in Google Street View images. The \textit{YoutubeDataset} dataset is gained from Youtube consisting of about ten million images~\cite{le2013building} and used in~\cite{szegedy2013intriguing}.

\subsection{Adversarial Examples and Countermeasures in Machine Learning}
Adversarial examples in conventional machine learning models have been discussed since decades ago. Machine learning-based systems with handcrafted features are primary targets, such as spam filters, intrusion detection, biometric authentication, fraud detection, etc.~\cite{barreno2010security}. For example, spam emails are often modified by adding characters to avoid detection~\cite{dalvi2004adversarial,lowd2005adversarial,biggio2010multiple}. 

Dalvi \etal first discussed adversarial examples and formulated this problem as a game between adversary and classifier (Na\"ive Bayes), both of which are sensitive to cost~\cite{dalvi2004adversarial}. The attack and defense on adversarial examples became an iterative game. Biggio \etal first tried a gradient-based approach to generate adversarial examples against linear classifier, support vector machine (SVM), and a neural network~\cite{biggio2013evasion}. Compared with deep learning adversarial examples, their methods allow more freedom to modify the data. The MNIST dataset was first evaluated under their attack, although a human could easily distinguish the adversarial digit images. Biggio \etal also reviewed several proactive defenses and discussed reactive approaches to improve the security of machine learning models~\cite{roli2013pattern}.


Barreno \etal presented an initial investigation on the security problems of machine learning~\cite{barreno2006can, barreno2010security}. They categorized attacking against machine learning system into three axes: 1) influence: whether attacks can poison the training data; 2) security violation: whether an adversarial example belongs to false positive or false negative; 3) specificity: attack is targeted to a particular instance or a wide class. We discuss these axes for deep learning area in Section~\ref{sec:taxonomy}. Barreno \etal compared attacks against SpamBayes spam filter and defenses as a study case. However, they mainly focused on binary classification problem such as virus detection system, intrusion detection system (IDS), and intrusion prevention system (IPS). 

Adversarial examples in conventional machine learning require knowledge of feature extraction, while deep learning usually needs only raw data input. In conventional ML, both attacking and defending methods paid great attention to features, even the previous step (data collection), giving less attention to the impact of humans. Then the target becomes a fully automatic machine learning system. Inspired by these studies on conventional ML, in this paper, we review recent security issues in the deep learning area. 

\cite{papernot2016towards} provided a comprehensive overview of security issues in machine learning and recent findings in deep learning. \cite{papernot2016towards} established a unifying threat model. A ``no free lunch'' theorem was introduced: the tradeoff between accuracy and robustness. 

Compared to their work, our paper focuses on adversarial examples in deep learning and has a detailed discussion on recent studies and findings.

For example, adversarial examples in an image classification task can be described as follows: Using a trained image classifier published by a third party, a user inputs one image to get the prediction of class label. Adversarial images are original clean images with small perturbations, often barely recognizable by humans. However, such perturbations misguide the image classifier. The user will get a response of an incorrect image label. Given a trained deep learning model $f$ and an original input data sample $x$, generating an adversarial example $x'$ can generally be described as a box-constrained optimization problem:
\begin{equation}
\begin{aligned}
\label{eq:general}
&\min_{x'} & & \|x'-x\| \\
&s.t. & & f(x') =l', \\
& & & f(x) = l, \\
& & & l \neq l', \\
& & & x' \in [0, 1],
\end{aligned}
\end{equation}
where $l$ and $l'$ denote the output label of $x$ and $x'$, $\|\cdot\|$ denotes the distance between two data sample. Let $\eta = x' - x$ be the perturbation added on $x$. This optimization problem minimizes the perturbation while misclassifying the prediction with a constraint of input data. 
In the rest of the paper, we will discuss variants of generating adversarial images and adversarial examples in other tasks.

%


\section{Taxonomy of Adversarial Examples}
\label{sec:taxonomy}
To systematically analyze approaches for generating adversarial examples, we analyze the approaches for generating adversarial examples (see details in Section~\ref{sec:method}) and categorize them along three dimensions: threat model, perturbation, and benchmark. 

\subsection{Threat Model}
We discuss the threat model in Section~\ref{sec:intro}. Based on different scenarios, assumptions, and quality requirements, adversaries decide the attributes they need in adversarial examples and then deploy specific attack approaches. We further decompose the threat model into four aspects: adversarial falsification, adversary's knowledge, adversarial specificity, and attack frequency. For example, if an adversarial example is required to be generated in real-time, adversaries should choose a one-time attack instead of an iterative attack, in order to complete the task (see Attack Frequency).

\begin{itemize}[leftmargin=1em]
\item Adversarial Falsification
\begin{itemize}
\item \textit{False positive} attacks generate a negative sample which is misclassified as a positive one (Type I Error). In a malware detection task, a benign software being classified as malware is a false positive. In an image classification task, a false positive can be an adversarial image unrecognizable to human, while deep neural networks predict it to a class with a high confidence score.  Figure~\ref{fig:unrecognizable} illustrates a false positive example of image classification.
\item \textit{False negative} attacks generate a positive sample which is misclassified as a negative one (Type II Error). In a malware detection task, a false negative can be the condition that a malware (usually considered as positive) cannot be identified by the trained model. False negative attack is also called machine learning evasion. This error is shown in most adversarial images, where human can recognize the image, but the neural networks cannot identify it. 

\end{itemize}

\item Adversary's Knowledge
\begin{itemize}[leftmargin=1em]
\item \textit{White-box} attacks assume the adversary knows everything related to trained neural network models, including training data, model architectures, hyper-parameters, numbers of layers, activation functions, model weights. Many adversarial examples are generated by calculating model gradients. Since deep neural networks tend to require only raw input data without handcrafted features and to deploy end-to-end structure, feature selection is not necessary compared to adversarial examples in machine learning. 
\item \textit{Black-box} attacks assume the adversary has no access to the trained neural network model. The adversary, acting as a standard user, only knows the output of the model (label or confidence score). 
This assumption is common for attacking online Machine Learning services (\eg, Machine Learning on AWS\footnote{\url{https://aws.amazon.com/machine-learning}}, Google Cloud AI\footnote{\url{https://cloud.google.com/products/machine-learning}}, BigML\footnote{\url{https://bigml.com}}, Clarifai\footnote{\url{https://www.clarifai.com}}, Microsoft Azure\footnote{\url{https://azure.microsoft.com/en-us/services/machine-learning}}, IBM Bluemix\footnote{\url{https://console.bluemix.net/catalog/services/machine-learning}}, Face++\footnote{\url{https://www.faceplusplus.com}}).

Most adversarial example attacks are \textit{white-box attacks}. However, they can be transferred to attack \textit{black-box} services due to the transferability of adversarial examples proposed by Papernot \etal~\cite{papernot2016transferability}. We will elaborate on it in Section~\ref{sec:transfer}. 
\end{itemize}

\item Adversarial Specificity
\begin{itemize}[leftmargin=1em]
\item \textit{Targeted} attacks misguide deep neural networks to a specific class. Targeted attacks usually occur in the multi-class classification problem. For example, an adversary fools an image classifier to predict all adversarial examples as one class. In a face recognition/biometric system, an adversary tries to disguise a face as an authorized user (Impersonation)~\cite{sharif2016accessorize}. Targeted attacks usually maximize the probability of targeted adversarial class.
\item \textit{Non-targeted} attacks do not assign a specific class to the neural network output. The adversarial class of output can be arbitrary except the original one. For example, an adversary makes his/her face misidentified as an arbitrary face in face recognition system to evade detection (dodging)~\cite{sharif2016accessorize}. 
Non-targeted attacks are easier to implement compared to \textit{targeted} attacks since it has more options and space to redirect the output. Non-targeted adversarial examples are usually generated in two ways: 1) running several targeted attacks and taking the one with the smallest perturbation from the results; 2) minimizing the probability of the correct class.

Some generation approaches (\eg, extended BIM, ZOO) can be applied to both targeted and non-targeted attacks. For binary classification, \textit{targeted} attacks are equivalent to \textit{non-targeted} attacks.
\end{itemize}

\item Attack Frequency
\begin{itemize}[leftmargin=1em]
\item \textit{One-time attacks} take only one time to optimize the adversarial examples. 
\item \textit{Iterative attacks} take multiple times to update the adversarial examples. 

Compared with \textit{one-time attacks}, \textit{iterative attacks} usually perform better adversarial examples, but require more interactions with victim classifier (more queries) and cost more computational time to generate them. For some computational-intensive tasks (\eg, reinforcement learning), \textit{one-time attacking} may be the only feasible choice.
\end{itemize}
\end{itemize}

\subsection{Perturbation}
Small perturbation is a fundamental premise for adversarial examples. Adversarial examples are designed to be close to the original samples and imperceptible to a human, which causes the performance degradation of deep learning models compared to that of a human. We analyze three aspects of perturbation: perturbation scope, perturbation limitation, and perturbation measurement.

\begin{itemize}[leftmargin=1em]
\item Perturbation Scope
\begin{itemize}
\item \textit{Individual} attacks generate different perturbations for each clean input.
\item \textit{Universal} attacks only create a universal perturbation for the whole dataset. This perturbation can be applied to all clean input data.

Most of the current attacks generate adversarial examples individually. However, \textit{universal} perturbations make it easier to deploy adversary examples in the real world. Adversaries do not require to change the perturbation when the input sample changes.
\end{itemize}

\item Perturbation Limitation
\begin{itemize}[leftmargin=1em]
\item \textit{Optimized Perturbation} sets perturbation as the goal of the optimization problem. These methods aim to minimize the perturbation so that humans cannot recognize the perturbation.
\item \textit{Constraint Perturbation} sets perturbation as the constraint of the optimization problem. These methods only require the perturbation to be small enough.
\end{itemize}

\item Perturbation Measurement
\begin{itemize}[leftmargin=1em]
\item \textit{$\ell_p$} measures the magnitude of perturbation by $p$-norm distance:
\begin{equation}
\|x\|_p = \left(\sum_{i=1}^n\|x_i\|^p\right)^{\frac{1}{p}}.
\end{equation}
$\ell_0, \ell_2, \ell_{\infty}$ are three commonly used $\ell_p$ metrics. $\ell_0$ counts the number of pixels changed in the adversarial examples; $\ell_2$ measures the Euclidean distance between the adversarial example and the original sample; $\ell_{\infty}$ denotes the maximum change for all pixels in adversarial examples.

\item \textit{Psychometric perceptual adversarial similarity score (PASS)} is a new metric introduced in~\cite{rozsa2016adversarial}, consistent with human perception.
\end{itemize}
\end{itemize}

\subsection{Benchmark}
Adversaries show the performance of their adversarial attacks based on different datasets and victim models. This inconsistency brings obstacles to evaluate the adversarial attacks and measure the robustness of deep learning models. Large and high-quality datasets, complex and high-performance deep learning models usually make adversaries/defenders hard to attack/defend. The diversity of datasets and victim models also makes researchers hard to tell whether the existence of adversarial examples is due to datasets or models.  We will discuss this problem in Section~\ref{sec:analysis_evaluation}. 

\begin{itemize}[leftmargin=1em]
\item Datasets\\
\textit{MNIST}, \textit{CIFAR-10}, and \textit{ImageNet} are three most widely used image classification datasets to evaluate adversarial attacks. Because \textit{MNIST} and \textit{CIFAR-10} are proved easy to attack and defend due to its simplicity and small size, \textit{ImageNet} is the best dataset to evaluate adversarial attacks so far. A well-designed dataset is required to evaluate adversarial attacks.

\item Victim Models\\
Adversaries usually attack several well-known deep learning models, such as \textit{LeNet}, \textit{VGG}, \textit{AlexNet}, \textit{GoogLeNet}, \textit{CaffeNet}, and \textit{ResNet}. 
\end{itemize}

In the following sections, we will investigate recent studies on adversarial examples according to this taxonomy. 

\section{Methods for generating Adversarial Examples}
\label{sec:method}
In this section, we illustrate several representative approaches for generating adversarial examples. Although many of these approaches are defeated by a countermeasure in later studies, we present these methods to show how the adversarial attacks improved and to what extent state-of-the-art adversarial attacks can achieve. The existence of these methods also requires investigation, which may improve the robustness of deep neural networks.

\begin{table*}[!t]
\centering
\caption{Taxonomy of Adversarial Examples}
\label{tab:taxonomy}
\begin{tabular}{|c|c|c|c|}
\hline
\multicolumn{1}{|c|}{\multirow{8}{*}{Threat Model}} & \multicolumn{1}{c|}{\multirow{2}{*}{Adversarial Falsification}} & \multicolumn{1}{c|}{False Negative}    & \multicolumn{1}{c|}{\cite{szegedy2013intriguing, goodfellow2014explaining, kurakin2016adversarial, papernot2016limitations, moosavi2016deepfool, carlini2017towards, chen2017zoo, moosavi2016universal, su2017one, sabour2016adversarial, rozsa2016adversarial, zhao2017generating, liu2017delving, carlini2017ground,tabacof2016exploring, kurakin2017scale, dong2017boosting}} \\ \cline{3-4} 
\multicolumn{1}{|c|}{}                              & \multicolumn{1}{c|}{}                                           & \multicolumn{1}{c|}{False Positive}    & \multicolumn{1}{c|}{\cite{nguyen2015deep}} \\ \cline{2-4} 
\multicolumn{1}{|c|}{}                              & \multicolumn{1}{c|}{\multirow{2}{*}{Adversary's Knowledge}}     & \multicolumn{1}{c|}{White-Box}         & \multicolumn{1}{c|}{\cite{szegedy2013intriguing, goodfellow2014explaining, kurakin2016adversarial, papernot2016limitations, moosavi2016deepfool, nguyen2015deep, carlini2017towards, moosavi2016universal, sabour2016adversarial, rozsa2016adversarial, liu2017delving, carlini2017ground, tabacof2016exploring, kurakin2017scale, dong2017boosting}} \\ \cline{3-4} 
\multicolumn{1}{|c|}{}                              & \multicolumn{1}{c|}{}                                           & \multicolumn{1}{c|}{Black-Box}         & \multicolumn{1}{c|}{\cite{chen2017zoo, su2017one, zhao2017generating, dong2017boosting}} \\ \cline{2-4} 
\multicolumn{1}{|c|}{}                              & \multicolumn{1}{c|}{\multirow{2}{*}{Adversarial Specificity}}   & \multicolumn{1}{c|}{Targeted}          & \multicolumn{1}{c|}{\cite{szegedy2013intriguing, papernot2016limitations, carlini2017towards, chen2017zoo, su2017one, sabour2016adversarial, rozsa2016adversarial, liu2017delving, carlini2017ground, tabacof2016exploring, kurakin2017scale, dong2017boosting}} \\ \cline{3-4} 
\multicolumn{1}{|c|}{}                              & \multicolumn{1}{c|}{}                                           & \multicolumn{1}{c|}{Non-Targeted}      & \multicolumn{1}{c|}{\cite{goodfellow2014explaining,kurakin2016adversarial, moosavi2016deepfool, nguyen2015deep, chen2017zoo, moosavi2016universal, su2017one, zhao2017generating, liu2017delving, dong2017boosting}} \\ \cline{2-4} 
\multicolumn{1}{|c|}{}                              & \multicolumn{1}{c|}{\multirow{2}{*}{Attack Frequency}}          & \multicolumn{1}{c|}{One-time}          & \multicolumn{1}{c|}{\cite{goodfellow2014explaining, rozsa2016adversarial}} \\ \cline{3-4} 
\multicolumn{1}{|c|}{}                              & \multicolumn{1}{c|}{}                                           & \multicolumn{1}{c|}{Iterative}         & \multicolumn{1}{c|}{\cite{szegedy2013intriguing, kurakin2016adversarial, papernot2016limitations, moosavi2016deepfool, nguyen2015deep, carlini2017towards, chen2017zoo, moosavi2016universal, su2017one, sabour2016adversarial, zhao2017generating, liu2017delving, carlini2017ground, tabacof2016exploring, kurakin2017scale, dong2017boosting}} \\ \hline
\multicolumn{1}{|c|}{\multirow{11}{*}{Perturbation}} & \multicolumn{1}{c|}{\multirow{2}{*}{Perturbation Scope}}        & \multicolumn{1}{c|}{Individual}        & \multicolumn{1}{c|}{\cite{szegedy2013intriguing, goodfellow2014explaining,kurakin2016adversarial, papernot2016limitations, moosavi2016deepfool, nguyen2015deep, carlini2017towards, chen2017zoo, su2017one, sabour2016adversarial, rozsa2016adversarial, zhao2017generating, liu2017delving, carlini2017ground, tabacof2016exploring, dong2017boosting}} \\ \cline{3-4} 
\multicolumn{1}{|c|}{}                              & \multicolumn{1}{c|}{}                                           & \multicolumn{1}{c|}{Universal}         & \multicolumn{1}{c|}{\cite{moosavi2016universal}} \\ \cline{2-4} 
\multicolumn{1}{|c|}{}                              & \multicolumn{1}{c|}{\multirow{3}{*}{Perturbation Limitation}}   & \multicolumn{1}{c|}{Optimized}         & \multicolumn{1}{c|}{\cite{szegedy2013intriguing, papernot2016limitations, moosavi2016deepfool, carlini2017towards, chen2017zoo, moosavi2016universal, rozsa2016adversarial, zhao2017generating, carlini2017ground, tabacof2016exploring}} \\ \cline{3-4} 
\multicolumn{1}{|c|}{}                              & \multicolumn{1}{c|}{}                                           & \multicolumn{1}{c|}{Constraint}        & \multicolumn{1}{c|}{\cite{su2017one, sabour2016adversarial, liu2017delving,rozsa2016adversarial, dong2017boosting}} \\ \cline{3-4} 
\multicolumn{1}{|c|}{}                              & \multicolumn{1}{c|}{}                                           & \multicolumn{1}{c|}{None}              & \multicolumn{1}{c|}{\cite{goodfellow2014explaining, kurakin2016adversarial, nguyen2015deep, kurakin2017scale}} \\ \cline{2-4} 
\multicolumn{1}{|c|}{}                              & \multicolumn{1}{c|}{\multirow{7}{*}{Perturbation Measurement}}  & \multicolumn{1}{c|}{Element-wise}      & \multicolumn{1}{c|}{\cite{goodfellow2014explaining, kurakin2016adversarial, kurakin2017scale}} \\ \cline{3-4} 
\multicolumn{1}{|c|}{}                              & \multicolumn{1}{c|}{}                                           & \multicolumn{1}{c|}{\multirow{4}{*}{$\ell_p (p\geq0)$}} & \multicolumn{1}{c|}{$\ell_0$: \cite{carlini2017towards, su2017one},} \\
\multicolumn{1}{|c|}{}                              & \multicolumn{1}{c|}{}                                           & \multicolumn{1}{c|}{} & 
\multicolumn{1}{c|}{$\ell_1$: \cite{carlini2017ground},} \\
\multicolumn{1}{|c|}{}                              & \multicolumn{1}{c|}{}                                           & \multicolumn{1}{c|}{} & 
\multicolumn{1}{c|}{$\ell_2$: \cite{moosavi2016deepfool, moosavi2016universal, szegedy2013intriguing, papernot2016limitations, carlini2017towards, chen2017zoo, sabour2016adversarial, zhao2017generating, liu2017delving, tabacof2016exploring, dong2017boosting},} \\
\multicolumn{1}{|c|}{}                              & \multicolumn{1}{c|}{}                                           & \multicolumn{1}{c|}{} & \multicolumn{1}{c|}{$\ell_{\infty}$: \cite{moosavi2016deepfool, moosavi2016universal, carlini2017towards, carlini2017ground, dong2017boosting},} \\ \cline{3-4} 
\multicolumn{1}{|c|}{}                              & \multicolumn{1}{c|}{}                                           & \multicolumn{1}{c|}{PASS}              & 
\multicolumn{1}{c|}{\cite{rozsa2016adversarial}} \\ \cline{3-4} 
\multicolumn{1}{|c|}{}                              & \multicolumn{1}{c|}{}                                           & \multicolumn{1}{c|}{None}              & \multicolumn{1}{c|}{\cite{nguyen2015deep}} \\ \hline
\multicolumn{1}{|c|}{\multirow{13}{*}{Benchmark}}                     & \multicolumn{1}{c|}{\multirow{5}{*}{Datasets}}                  & \multicolumn{1}{c|}{MNIST}             & \multicolumn{1}{c|}{\cite{szegedy2013intriguing, goodfellow2014explaining, papernot2016limitations, moosavi2016deepfool, nguyen2015deep, carlini2017towards, rozsa2016adversarial, zhao2017generating, carlini2017ground, tabacof2016exploring}} \\ \cline{3-4} 
\multicolumn{1}{|c|}{}                              & \multicolumn{1}{c|}{}                                           & \multicolumn{1}{c|}{CIFAR-10}           & \multicolumn{1}{c|}{\cite{moosavi2016deepfool, carlini2017towards, chen2017zoo, su2017one}} \\  \cline{3-4} 
\multicolumn{1}{|c|}{}                              & \multicolumn{1}{c|}{}                                           & \multicolumn{1}{c|}{ImageNet}          & \multicolumn{1}{c|}{\cite{szegedy2013intriguing, goodfellow2014explaining, kurakin2016adversarial, moosavi2016deepfool, nguyen2015deep, carlini2017towards, chen2017zoo, moosavi2016universal, sabour2016adversarial, rozsa2016adversarial, liu2017delving, tabacof2016exploring, kurakin2017scale, dong2017boosting}} \\  \cline{3-4} 
\multicolumn{1}{|c|}{}                              & \multicolumn{1}{c|}{}                                           & \multicolumn{1}{c|}{\multirow{2}{*}{Others}}            &  \multicolumn{1}{c|}{YoutubeDataset: \cite{szegedy2013intriguing}} \\ 
\multicolumn{1}{|c|}{}                              & \multicolumn{1}{c|}{}                                           & \multicolumn{1}{c|}{}            & \multicolumn{1}{c|}{LSUN, SNLI: \cite{zhao2017generating}} \\ \cline{2-4}
\multicolumn{1}{|c|}{}                              & \multicolumn{1}{c|}{\multirow{8}{*}{Victim Models}}                                     & \multicolumn{1}{c|}{LeNet}             & \multicolumn{1}{c|}{\cite{papernot2016limitations, moosavi2016deepfool, nguyen2015deep, rozsa2016adversarial, zhao2017generating}} \\ \cline{3-4} 
\multicolumn{1}{|c|}{}                              & \multicolumn{1}{c|}{}                                           & \multicolumn{1}{c|}{VGG}               & \multicolumn{1}{c|}{\cite{moosavi2016universal, su2017one, sabour2016adversarial, liu2017delving}} \\ \cline{3-4} 
\multicolumn{1}{|c|}{}                              & \multicolumn{1}{c|}{}                                           & \multicolumn{1}{c|}{AlexNet}           & \multicolumn{1}{c|}{\cite{nguyen2015deep, sabour2016adversarial}} \\ \cline{3-4} 
\multicolumn{1}{|c|}{}                              & \multicolumn{1}{c|}{}                                           & \multicolumn{1}{c|}{QuocNet}           & \multicolumn{1}{c|}{\cite{szegedy2013intriguing}} \\ \cline{3-4} 
\multicolumn{1}{|c|}{}                              & \multicolumn{1}{c|}{}                                           & \multicolumn{1}{c|}{GoogLeNet}         & \multicolumn{1}{c|}{ \cite{goodfellow2014explaining, kurakin2016adversarial, moosavi2016deepfool, carlini2017towards, chen2017zoo, moosavi2016universal, sabour2016adversarial, rozsa2016adversarial, zhao2017generating, liu2017delving, kurakin2017scale, dong2017boosting}} \\ \cline{3-4} 
\multicolumn{1}{|c|}{}                              & \multicolumn{1}{c|}{}                                           & \multicolumn{1}{c|}{CaffeNet}          & \multicolumn{1}{c|}{\cite{moosavi2016deepfool, moosavi2016universal, sabour2016adversarial}} \\ \cline{3-4} 
\multicolumn{1}{|c|}{}                              & \multicolumn{1}{c|}{}                                           & \multicolumn{1}{c|}{ResNet}            & \multicolumn{1}{c|}{\cite{moosavi2016universal, rozsa2016adversarial, liu2017delving, dong2017boosting}} \\ \cline{3-4} 
\multicolumn{1}{|c|}{}                              & \multicolumn{1}{c|}{}                                           & \multicolumn{1}{c|}{LSTM}              & \multicolumn{1}{c|}{\cite{zhao2017generating}} \\ \hline 
\end{tabular}
\end{table*}

Table~\ref{tab:taxonomy} summaries the methods for generating adversarial examples in this section based on the proposed taxonomy. 

\subsection{L-BFGS Attack}
Szegedy \etal first introduced adversarial examples against deep neural networks in 2014~\cite{szegedy2013intriguing}. They generated adversarial examples using a L-BFGS method to solve the general targeted problem:
\begin{equation}
\begin{aligned}
&\min_{x'} && c\|\eta\|+J_{\theta}(x', l') \\
&s.t. && x'\in [0, 1].
\end{aligned}
\end{equation}

To find a suitable constant $c$, \textit{L-BFGS Attack} calculated approximate values of adversarial examples by line-searching $c>0$. The authors showed that the generated adversarial examples could also be generalized to different models and different training datasets. 
They suggested that adversarial examples are never/rarely seen examples in the test datasets.

\textit{L-BFGS Attack} was also used in~\cite{tabacof2016exploring}, which implemented a binary search to find the optimal $c$.

\subsection{Fast Gradient Sign Method (FGSM)}
\begin{figure}[!t]
\centering
\includegraphics[width=0.45\textwidth]{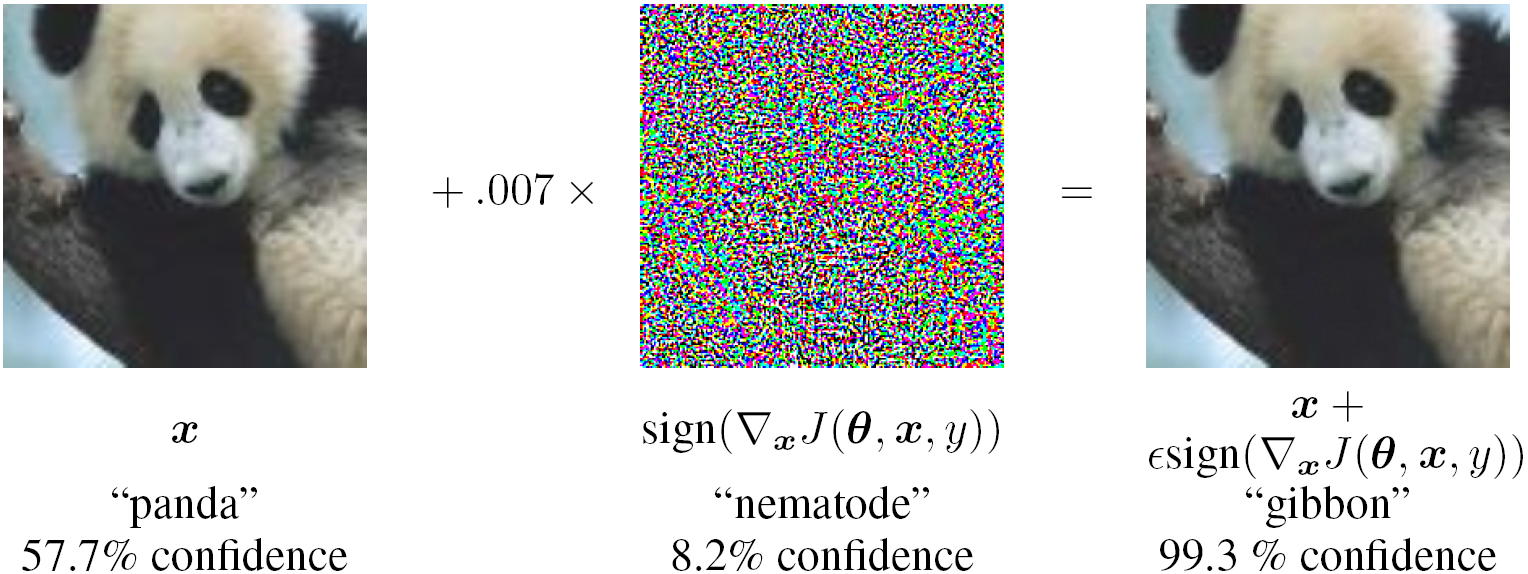}
\caption{An adversarial image generated by \textit{Fast Gradient Sign Method}~\cite{goodfellow2014explaining}: left: a clean image of a panda; middle: the perturbation; right: one sample adversarial image, classified as a gibbon.}
\label{fig:panda}
\vspace{-15pt}
\end{figure}

\textit{L-BFGS Attack} used an expensive linear search method to find the optimal value, which was time-consuming and impractical. Goodfellow \etal proposed a fast method called \textit{Fast Gradient Sign Method} to generate adversarial examples~\cite{goodfellow2014explaining}. They only performed one step gradient update along the direction of the sign of gradient at each pixel. Their perturbation can be expressed as:
\begin{equation}
\eta = \epsilon sign (\nabla_x J_{\theta}(x, l)),
\end{equation}
where $\epsilon$ is the magnitude of the perturbation. The generated adversarial example $x'$ is calculated as: $x'=x+\eta$. This perturbation can be computed by using back-propagation. Figure~\ref{fig:panda} shows an adversarial example on ImageNet.

They claimed that the linear part of the high dimensional deep neural network could not resist adversarial examples, although the linear behavior speeded up training. Regularization approaches are used in deep neural networks such as dropout. Pre-training could not improve the robustness of networks. 

\cite{rozsa2016adversarial} proposed a new method, called \textit{Fast Gradient Value} method, in which they replaced the sign of the gradient with the raw gradient: $\eta = \nabla_x J(\theta, x, l)$. \textit{Fast Gradient Value} method has no constraints on each pixel and can generate images with a larger local difference.

According to~\cite{kurakin2017scale}, one-step attack is easy to transfer but also easy to defend (see Section~\ref{sec:transfer}). \cite{dong2017boosting} applied momentum to \textit{FGSM} to generate adversarial examples more iteratively. The gradients were calculated by:
\begin{equation}
\begin{aligned}
\mathbf{g}_{t+1} & = \mu \mathbf{g}_t + \frac{\nabla_x J_{\theta}(x'_t, l)}{\|\nabla_x J_{\theta}(x'_t, l)\|},
\end{aligned}
\end{equation}
then the adversarial example is derived by $x'_{t+1} = x'_t+\epsilon sign{\mathbf{g}_{t+1}}$. The authors increased the effectiveness of attack by introducing momentum and improved the transferability by applying the one-step attack and the ensembling method. 

\cite{kurakin2017scale} extended \textit{FGSM} to a targeted attack by maximizing the probability of the target class:
\begin{equation}
    x' = x - \epsilon sign(\nabla_x J(\theta, x, l')).
\end{equation}
The authors refer to this attack as \textit{One-step Target Class Method (OTCM)}.

\cite{tramer2017ensemble} found that \textit{FGSM} with adversarial training is more robust to white-box attacks than to black-box attacks due to \textit{gradient masking}. They proposed a new attack, \textit{RAND-FGSM}, which added random when updating the adversarial examples to defeat adversarial training:
\begin{equation}
\begin{aligned}
x_{tmp} &= x+\alpha \cdot sign(\mathcal{N}(\mathbf{0}^d, \mathbf{I}^d)),\\
x' &= x_{tmp} + (\epsilon-\alpha) \cdot sign(\nabla_{x_{tmp}}J(x_{tmp}, l)),
\end{aligned}
\end{equation}
where $\alpha, \epsilon$ are the parameters, $\alpha < \epsilon$.

\subsection{Basic Iterative Method (BIM) and Iterative Least-Likely Class Method (ILLC)}
Previous methods assume adversarial data can be directly fed into deep neural networks. However, in many applications, people can only pass data through devices (\eg, cameras, sensors). Kurakin \etal applied adversarial examples to the physical world~\cite{kurakin2016adversarial}. They extended \textit{Fast Gradient Sign} method by running a finer optimization (smaller change) for multiple iterations. In each iteration, they clipped pixel values to avoid large change on each pixel:
\begin{equation}
Clip_{x, \xi}\{x'\} = \min\{255, x+\xi, \max\{0, x-\epsilon, x'\}\},
\end{equation}
where $Clip_{x, \xi}\{x'\}$ limits the change of the generated adversarial image in each iteration. The adversarial examples were generated in multiple iterations:
\begin{equation}
\begin{aligned}
x_0 &= x, \\
x_{n+1} &= Clip_{x, \xi}\{x_n + \epsilon sign(\nabla_x J(x_n, y))\}.
\end{aligned}
\end{equation}
The authors referred to this method as \textit{Basic Iterative} method.

To further attack a specific class, they chose the least-likely class of the prediction and tried to maximize the cross-entropy loss. This method is referred to as \textit{Iterative Least-Likely Class} method:
\begin{equation}
\begin{aligned}
x_0 &= x, \\
y_{LL} &= {arg\min}_y\{p(y|x)\}, \\
x_{n+1} &= Clip_{x, \epsilon}\{x_n - \epsilon sign(\nabla_xJ(x_n, y_{LL}))\}.
\end{aligned}
\end{equation}
They successfully fooled the neural network with a crafted image taken from a cellphone camera. They also found that \textit{Fast Gradient Sign} method is robust to phototransformation, while iterative methods cannot resist phototransformation.

\subsection{Jacobian-based Saliency Map Attack (JSMA)}
Papernot \etal designed an efficient saliency adversarial map, called \textit{Jacobian-based Saliency Map Attack}~\cite{papernot2016limitations}. They first computed Jacobian matrix of given sample $x$, which is given by:
\begin{equation}
J_F(x) = \frac{\partial F(x)}{\partial x} = \bigg[\frac{\partial F_j(x)}{\partial x_i}\bigg]_{i\times j}.
\end{equation}
According to~\cite{carlini2017towards}, $F$ denotes the second-to-last layer (logits) in~\cite{papernot2016limitations}. Carlini and Wagner modify this approach by using the output of the softmax layer as $F$~\cite{carlini2017towards}.
In this way, they found the input features of $x$ that made most significant changes to the output. A small perturbation was designed to successfully induce large output variations so that change in a small portion of features could fool the neural network. 

Then the authors defined two adversarial saliency maps to select the feature/pixel to be crafted in each iteration. 
They achieved 97\% adversarial success rate by modifying only 4.02\% input features per sample. However, this method runs very slow due to its significant computational cost.

\subsection{DeepFool}
Moosavi-Dezfooli \etal proposed \textit{DeepFool} to find the closest distance from the original input to the decision boundary of adversarial examples~\cite{moosavi2016deepfool}. To overcome the non-linearity in high dimension, they performed an iterative attack with a linear approximation. Starting from an affine classifier, they found that the minimal perturbation of an affine classifier is the distance to the separating affine hyperplane $\mathcal{F}=\{x: w^Tx+b=0\}$ 
. The perturbation of an affine classifier $f$ can be $\eta^*(x) = -\frac{f(x)}{\|w\|^2}w$.

If $f$ is a binary differentiable classifier, they used an iterative method to approximate the perturbation by considering $f$ is linearized around $x_i$ at each iteration. The minimal perturbation is computed as:
\begin{equation}
\begin{aligned}
&\underset{\eta_i}{\arg\min} && \|\eta_i\|_2  \\
&s.t. && f(x_i) + \nabla f(x_i)^T \eta_i = 0.
\end{aligned}
\end{equation}
This result can also be extended to the multi-class classifier by finding the closest hyperplanes. It can also be extended to a more general $\ell_p$ norm, $p\in [0, \infty)$. \textit{DeepFool} provided less perturbation compared to \textit{FGSM} and \textit{JSMA} did. Compared to \textit{JSMA}, \textit{DeepFool} also reduced the intensity of perturbation instead of the number of selected features.

\subsection{CPPN EA Fool}
\begin{figure}[!t]
\centering
\includegraphics[width=0.9\linewidth]{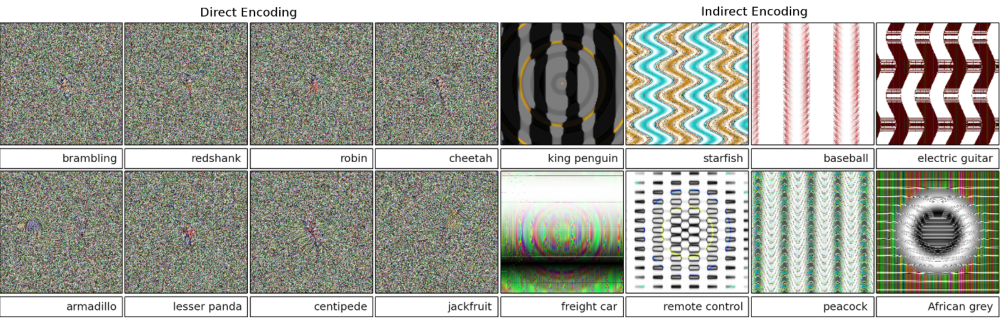}
\caption{Unrecognizable examples to humans, but deep neural networks classify them to a class with high certainty ($\geq 99.6\%$)~\cite{nguyen2015deep}} 
\label{fig:unrecognizable}
\vspace{-15pt}
\end{figure}

Nguyen \etal discovered a new type of attack, compositional pattern-producing network-encoded EA (CPPN EA),  where adversarial examples are classified by deep neural networks with high confidence (99\%), which is unrecognizable to human~\cite{nguyen2015deep}. We categorize this kind of attack as a \textit{False positive} attack. Figure~\ref{fig:unrecognizable} illustrates false-positive adversarial examples.

They used evolutionary algorithms (EAs) algorithm to produce the adversarial examples. To solve multi-class classification problem using EA algorithms, they applied multi-dimensional archive of phenotypic elites MAP-Elites~\cite{cully2015robots}.  The authors first encoded images with two different methods: direct encoding (grayscale or HSV value) and indirect encoding (compositional pattern-producing network). Then in each iteration, MAP-Elites, like general EA algorithm, chose a random organism, mutated them randomly, and replaced with the current ones if the new ones have higher \textit{fitness} (high certainty for a class of a neural network). In this way, MAP-Elites can find the best individual for each class. As they claimed, for many adversarial images, CPPN could locate the critical features to change outputs of deep neural networks just like JSMA did. Many images from same evolutionary are found similar on closely related categories. More interestingly, CPPN EA fooling images are accepted by an art contest with 35.5\% acceptance rate.



\subsection{C\&W's Attack}
Carlini and Wagner launched a targeted attack to defeat \textit{Defensive distillation} (Section~\ref{sec:distillation})~\cite{carlini2017towards}. According to their further study~\cite{carlini2017adversarial,carlini2017magnet}, \textit{C\&W's Attack} is effective for most of existing adversarial detecting defenses. The authors made several modifications in Equation~\ref{eq:general}. 

They first defined a new objective function $g$, so that: 
\begin{equation}
\begin{aligned}
&\min_\eta && \|\eta\|_p + c\cdot g(x+\eta) \\
&s.t. &&  x+\eta \in [0,1]^n,
\end{aligned}
\end{equation}
where $g(x')\geq 0$ if and only if $f(x')=l'$. In this way, the distance and the penalty term can be better optimized. The authors listed seven objective function candidates $g$. One of the effective functions evaluated by their experiments can be: 
\begin{equation}
    g(x') = \max(\max_{i\neq l'}(Z(x')_i)-Z(x')_t, -\kappa),
\end{equation}
where $Z$ denotes the Softmax function, $\kappa$ is a constant to control the confidence ($\kappa$ is set to 0 in~\cite{carlini2017towards}).

Second, instead of using box-constrained L-BFGS to find minimal perturbation in \textit{L-BFGS Attack} method, the authors introduced a new variant $w$ to avoid the box constraint, where $w$ satisfies $\eta=\frac{1}{2}(\tanh (w)+1)-x$. General optimizers in deep learning like Adam and SGD were used to generate adversarial examples and performed 20 iterations of such generation to find an optimal $c$ by binary searching. However, they found that if the gradients of $\|\eta\|_p$ and $g(x+\eta)$ are not in the same scale, it is hard to find a suitable constant $c$ in all of the iterations of the gradient search and get the optimal result. Due to this reason, two of their proposed functions did not find optimal solutions for adversarial examples.

Third, three distance measurements of perturbation were discussed in the paper: $\ell_0$, $\ell_2$, and $\ell_\infty$. The authors provided three kinds of attacks based on the distance metrics: $\ell_0$ attack, $\ell_2$ attack, and $\ell_\infty$ attack.

$\ell_2$ attack can be described by:
\begin{equation}
\min_w \|\frac{1}{2}(\tanh(w)+1)\|_2+c\cdot g(\frac{1}{2}\tanh(w)+1).
\end{equation}
The authors showed that the distillation network could not help defend $\ell_2$ attack. 

$\ell_0$ attack was conducted iteratively since $\ell_0$ is not differentiable. In each iteration, a few pixels are considered trivial for generating adversarial examples and removed. The importance of pixels is determined by the gradient of $\ell_2$ distance. The iteration stops if the remaining pixels can not generate an adversarial example.

$\ell_{\infty}$ attack was also an iterative attack, which replaced the $\ell_2$ term with a new penalty in each iteration:
\begin{equation}
    \min \quad c\cdot g(x+\eta) + \sum_i[(\eta_i-\tau)^+].
\end{equation}
For each iteration, they reduced $\tau$ by a factor of 0.9, if all $\eta_i < \tau$. $\ell_{\infty}$ attack considered $\tau$ as an estimation of $\ell_\infty$.

\subsection{Zeroth Order Optimization (ZOO)}
Different from gradient-based adversarial generating approaches, Chen \etal proposed a \textit{Zeroth Order Optimization (ZOO)} based attack~\cite{chen2017zoo}. Since this attack does not require gradients, it can be directly deployed in a black-box attack without model transferring. Inspired by~\cite{carlini2017towards}, the authors modified $g(\cdot)$ in \cite{carlini2017towards} as a new hinge-like loss function:
\begin{equation}
    g(x') = \max(\max_{i\neq l'} (\log [f(x)]_i) - \log [f(x)]_{l'}, -\kappa),
\end{equation}
and used symmetric difference quotient to estimate the gradient and Hessian:
\begin{equation}
\begin{aligned}
    \frac{\partial f(x)}{\partial x_i} \approx & \frac{f(x+he_i)-f(x-he_i)}{2h}, \\
    \frac{\partial^2 f(x)}{\partial x_{i}^2} \approx & \frac{f(x+he_i)-2f(x)+f(x-he_i)}{h^2} ,
\end{aligned}
\end{equation}
where $e_i$ denotes the standard basis vector with the $i$th component as 1, $h$ is a small constant.

Through employing the gradient estimation of gradient and Hessian, \textit{ZOO} does not need the access to the victim deep learning models. However, it requires expensive computation to query and estimate the gradients. The authors proposed ADAM like algorithms, \textit{ZOO-ADAM}, to randomly select a variable and update adversarial examples.  Experiments showed that \textit{ZOO} achieved the comparable performance as \textit{C\&W's Attack}.

\subsection{Universal Perturbation}
Leveraging their previous method on \textit{DeepFool}, Moosavi-Dezfooli \etal developed a universal adversarial attack~\cite{moosavi2016universal}. The problem they formulated is to find a universal perturbation vector satisfying 
\begin{equation}
\begin{aligned}
\|\eta\|_p & \leq \epsilon, \\
\mathcal{P}\big(x' \neq f(x)\big) & \geq 1-\delta.
\end{aligned}
\end{equation}
$\epsilon$ limits the size of universal perturbation, and $\delta$ controls the failure rate of all the adversarial samples. 

\begin{figure}[!t]
\centering
\includegraphics[width=0.5\linewidth]{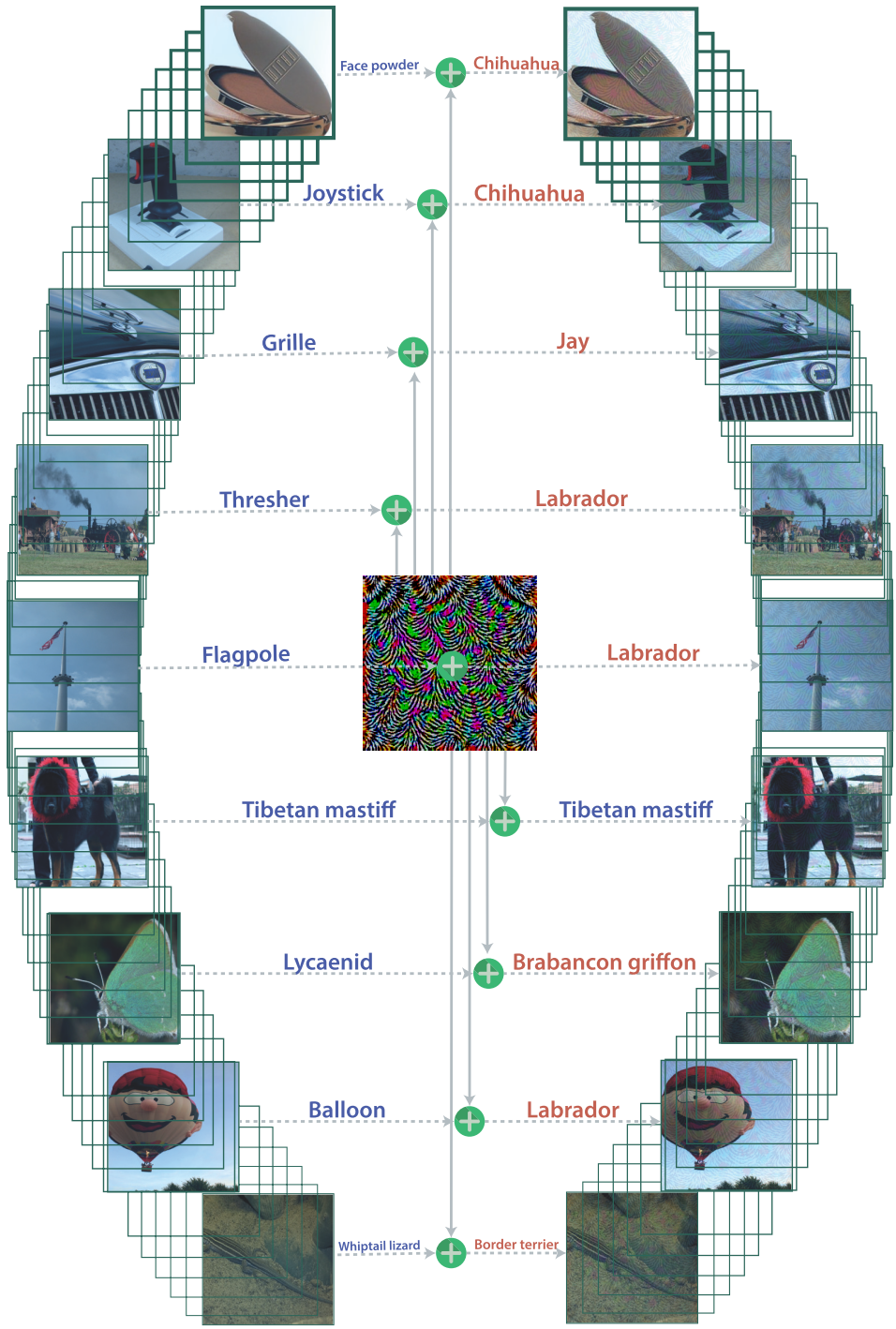}
\caption{A universal adversarial example fools the neural network on images. Left images: original labeled natural images; center image: universal perturbation; right images: perturbed images with wrong labels.~\cite{moosavi2016universal}}
\label{fig:universal}
\vspace{-15pt}
\end{figure}

For each iteration, they use \textit{DeepFool} method to get a minimal sample perturbation against each input data and update the perturbation to the total perturbation $\eta$. This loop will not stop until most data samples are fooled ($\mathcal{P}<1-\delta$). From experiments, the universal perturbation can be generated by using a small part of data samples instead of the entire dataset. 
Figure~\ref{fig:universal} illustrates a universal adversarial example can fool a group of images.
The universal perturbations were shown to be generalized well across popular deep learning architectures (\eg, VGG, CaffeNet, GoogLeNet, ResNet).


\subsection{One Pixel Attack}
To avoid the problem of measurement of perceptiveness, Su \etal generated adversarial examples by only modifying one pixel~\cite{su2017one}. The optimization problem becomes: 
\begin{equation}
\begin{aligned}
&\min_{x'} & & J(f(x'), l') \\
&s.t. & & \|\eta\|_0 \leq \epsilon_0, \\
\end{aligned}
\end{equation}
where $\epsilon_0=1$ for modifying only one pixel. The new constraint made it hard to optimize the problem. 

Su \etal applied differential evolution (DE), one of the evolutionary algorithms, to find the optimal solution. DE does not require the gradients of the neural networks and can be used in non-differential objective functions. They evaluated the proposed method on the \textit{CIFAR-10} dataset using three neural networks: All convolution network (AllConv)~\cite{springenberg2014striving}, Network in Network (NiN)~\cite{lin2013network}, and VGG16. Their results showed that 70.97\% of images successfully fooled deep neural networks with at least one target class with confidence 97.47\% on average.

\subsection{Feature Adversary}
Sabour \etal performed a targeted attack by minimizing the distance of the representation of internal neural network layers instead of the output layer~\cite{sabour2016adversarial}. We refer to this attack as \textit{Feature Adversary}. The problem can be described by:
\begin{equation}
\begin{aligned}
&\min_{x'} && \|\phi_k(x)-\phi_k(x')\| \\
&s.t. && \|x-x'\|_{\infty} < \delta,
\end{aligned}
\end{equation}
where $\phi_k$ denotes a mapping from image input to the output of the $k$th layer. Instead of finding a minimal perturbation, $\delta$ is used as a constraint of perturbation. They claimed that a small fixed value $\delta$ is good enough for human perception. Similar to~\cite{szegedy2013intriguing}, they used L-BFGS-B to solve the optimization problem. The adversarial images are more natural and closer to the targeted images in the internal layers. 


\subsection{Hot/Cold}
Rozsa \etal proposed a \textit{Hot/Cold} method to find multiple adversarial examples for every single image input~\cite{rozsa2016adversarial}. They thought small translations and rotations should be allowed as long as they were \textit{imperceptible}.  

They defined a new metric, Psychometric Perceptual Adversarial Similarity Score (PASS), to measure the noticeable similarity to humans. \textit{Hot/Cold} neglected the unnoticeable difference based on pixels and replaced widely used $\ell_p$ distance with PASS. PASS includes two stages: 1) aligning the modified image with the original image; 2) measuring the similarity between the aligned image and the original one. 

Let $\phi(x', x)$ be a homography transform from the adversarial example $x'$ to the original example $x$. $\mathcal{H}$ is the homography matrix, with size $3 \times 3$. $\mathcal{H}$ is solved by maximizing the \textit{enhanced correlation coefficient (ECC)}~\cite{evangelidis2008parametric} between $x'$ and $x$. The optimization function is:
\begin{equation}
\underset{\mathcal{H}}{\arg \min} \bigg\|\frac{\overline{x}}{\|\overline{x}\|}-\frac{\overline{\phi(x', x)}}{\|\overline{\phi(x', x)}\|}\bigg\|,
\end{equation}
where $\overline{\cdot}$ denotes the normalization of an image.

\textit{Structural SIMilarity (SSIM)} index~\cite{flynn2013image} was adopted to measure the just noticeable difference of images. \cite{rozsa2016adversarial} leveraged SSIM and defined a new measurement, regional SSIM index (RSSIM) as:
$$RSSIM(x_{i,j},x_{i,j}') = L(x_{i,j}, x_{i,j}')^{\alpha}C(x_{i,j}, x_{i,j}')^{\beta}S(x_{i,j}, x_{i,j}')^{\gamma},$$
where $\alpha, \beta, \gamma$ are weights of importance for luminance ($L(\cdot, \cdot)$), contrast ($C(\cdot, \cdot)$), and structure ($S(\cdot, \cdot)$).
The SSIM can be calculated by averaging RSSIM:
$$SSIM(x_{i,j},x_{i,j}') = \frac{1}{n\times m}\sum_{i,j} RSSIM(x_{i,j},x_{i,j}').$$

PASS is defined by combination of the alignment and the similarity measurement:
\begin{equation}
    PASS(x',x) = SSIM(\phi^* (x', x), x).
\end{equation}

The adversarial problem with the new distance is described as:
\begin{equation}
\begin{aligned}
&\min && D(x, x') \\
&s.t. && f(x') =  y', \\
&&& PASS(x,x') \geq \gamma. 
\end{aligned}
\end{equation}
$D(x, x')$ denotes a measure of distance (\eg, $1-PASS(x,x')$ or $\|x-x'\|_p$).

To generate a diverse set of adversarial examples, the authors defined the targeted label $l'$ as \textit{hot} class, and the original label $l$ as \textit{cold} class. In each iteration, they moved toward a target (hot) class while moving away from the original (cold) class. Their results showed that generated adversarial examples are comparable to \textit{FGSM}, and with more diversity.


\subsection{Natural GAN}
\label{sec:natural_gan}
Zhao \etal utilized Generative Adversarial Networks (GANs) as part of their approach to generate adversarial examples of images and texts~\cite{zhao2017generating}, which made adversarial examples more natural to human. We name this approach \textit{Natural GAN}. The authors first trained a WGAN model on the dataset, where the generator $\mathcal{G}$ maps random noise to the input domain. They also trained an ``inverter'' $\mathcal{I}$ to map input data to dense inner representations. Hence, the adversarial noise was generated by minimizing the distance of the inner representations like ``Feature Adversary.'' The adversarial examples were generated using the generator: $x' = \mathcal{G}(z')$:
\begin{equation}
\begin{aligned}
& \min_z && \|z-\mathcal{I}(x)\| \\
& s.t. && f(\mathcal{G}(z)) \neq f(x).
\end{aligned}
\end{equation}

Both the generator $\mathcal{G}$ and the inverter $\mathcal{I}$ were built to make adversarial examples natural. \textit{Natural GAN} was a general framework for many deep learning fields. \cite{zhao2017generating} applied \textit{Natural GAN} to image classification, textual entailment, and machine translation. Since \textit{Natural GAN} does not require gradients of original neural networks, it can also be applied to \textit{Black-box Attack}.


\subsection{Model-based Ensembling Attack}
Liu \etal conducted a study of transferability (Section~\ref{sec:transfer}) over deep neural networks on ImageNet and proposed a \textit{Model-based Ensembling Attack} for targeted adversarial examples~\cite{liu2017delving}. The authors argued that compared to non-targeted adversarial examples, targeted adversarial examples are much harder to transfer over deep models. Using \textit{Model-based Ensembling Attack}, they can generate transferable adversarial examples to attack a black-box model. 

The authors generated adversarial examples on multiple deep neural networks with full knowledge and tested them on a black-box model. \textit{Model-based Ensembling Attack} was derived by the following optimization problem:
\begin{equation}
\underset{x'}{\arg\min} \quad -\log\bigg((\sum_{i=1}^k\alpha_i J_i(x', l'))\bigg)+\lambda \|x'-x\|,
\end{equation}
where $k$ is the number of deep neural networks in the generation, $f_i$ is the function of each network, and $\alpha_i$ is the ensemble weight ($\sum_i^k \alpha_i = 1$). The results showed that \textit{Model-based Ensembling Attack} could generate transferable targeted adversarial images, which enhanced the power of adversarial examples for black-box attacks. They also proved that this method performs better in generating non-targeted adversarial examples than previous methods. The authors successfully conducted a black-box attack against \url{Clarifai.com} using \textit{Model-based Ensembling Attack}.

\subsection{Ground-Truth Attack}
Formal verification techniques aim to evaluate the robustness of a neural network even against zero-day attacks (Section~\ref{sec:verify}). Carlini \etal constructed a ground-truth attack, which provided adversarial examples with minimal perturbation ($\ell_1, \ell_\infty$)~\cite{carlini2017ground}. \textit{Network Verification} always checks whether an adversarial example violates a property of a deep neural network and whether there exists an example changes the label within a certain distance. \textit{Ground-Truth Attack} conducted a binary search and found such an adversarial example with the smallest perturbation by invoking Reluplex~\cite{katz2017reluplex} iteratively. The initial adversarial example is found using C\&W's Attack~\cite{carlini2017towards} to improve the performance.



\section{Applications for Adversarial Examples}
\label{sec:app}
We have investigated adversarial examples for image classification task. In this section, we review adversarial examples against the other tasks. We mainly focus on three questions: What scenarios are adversarial examples applied in new tasks? How to generate adversarial examples in new tasks? Whether to propose a new method or to translate the problem into the image classification task and solve it by the aforementioned methods? Table~\ref{tab:app} summarizes the applications for adversarial examples in this section.

\begin{table*}[!t]
\centering
\rotatebox{90}{
\begin{minipage}{0.97\textheight}
\caption{Summary of Applications for Adversarial Examples}
\label{tab:app}
\begin{tabularx}{0.97\textheight}{|X|Y|Y|Y|Y|Y|Y|Y|Y|Y|Y|Y|}
\hline
Applications                            & Representative Study             & Method                           & Adversarial Falsification & Adversary's Knowledge   & Adversarial Specificity         & Perturbation Scope & Perturbation Limitation & Attack Frequency    & Perturbation Measurement       & Dataset             & Architecture                                         \\ \hline
\multirow{2}{0.5in}{Reinforcement Learning} & \cite{huang2017adversarial}  & FGSM                             & N/A                              & {White-box \& Black-box} & Non-Targeted              & Individual         & N/A                     & One-time            & $\ell_1, \ell_2, \ell_\infty$ & Atari               & DQN, TRPO, A3C                                       \\ \cline{2-12} 
                                        & \cite{kos2017delving}    & FGSM                             & N/A                              & White-box               & Non-Targeted              & Individual         & N/A                     & One-time            & N/A                            & Atari Pong          & A3C                                                  \\ \hline
\multirow{2}{0.5in}{Generative Modeling}    & \cite{kos2017adversarial}          & Feature Adversary, C\&W          & N/A                              & White-box               & Targeted                  & Individual         & Optimized               & Iterative           & $\ell_2$                       & MNIST, SVHN, CelebA & VAE, VAE-GAN                                         \\ \cline{2-12} 
                                        & \cite{tabacof2016adversarial}      & Feature Adversary                & N/A                              & White-box               & Targeted                  & Individual         & Optimized               & Iterative           & $\ell_2$                       & MNIST, SVHN         & VAE, AE                                              \\ \hline
Face Recognition                        & \cite{sharif2016accessorize}                & Impersonation \& Dodging Attack & False negative                   & {white-box \& black-box} & {Targeted \& Non-Targeted} & Universal         & Optimized               & Iterative           & Total Variation                & LFW,                & VGGFace                                              \\ \hline
Object Detection                        & \cite{xie2017adversarial}          & DAG                              & False negative \& False positive & White-box \& Black-box & Non-Targeted              & Individual         & N/A                     & Iterative           & N/A                            & VOC2007, VOC2012    & Faster-RCNN                                          \\ \hline
\multirow{2}{0.5in}{Semantic Segmentation}  & \cite{xie2017adversarial}     & DAG                              & {False negative \& False positive} & {White-box \& Black-box} & Non-Targeted              & Individual         & N/A                     & Iterative           & N/A                            & DeepLab             & FCN                                                  \\ \cline{2-12} 
                                        & \cite{fischer2017adversarial} & ILLC                             & False negative                   & White-box               & Targeted                  & Individual         & N/A                     & Iterative           & $\ell_\infty$                 & Cityscapes          & FCN                                                  \\ \cline{2-12} 
                                        & \cite{metzen2017detecting}  & ILLC                             & False negative                   & White-box               & Targeted                  & Universal          & N/A                     & Iterative           & N/A                            & Cityscapes          & FCN                                                  \\ \hline
\multirow{2}{0.5in}{Reading Comprehension}                   & \cite{jia2017adversarial}     & AddSent, AddAny                  & N/A                              & Black-box               & Non-Targeted              & Individual         & N/A                     & One-time \& Iterative & N/A                            & SQuAD               & BiDAF, Match-LSTM, and twelve other published models \\ \cline{2-12}
& \cite{li2016understanding}    & Reinforcement Learning   & False negative    & White-box     & Non-Targeted      & Individual        & Optimized         & Iterative     & $\ell_0$  & TripAdvisor Dataset           & Bi-LSTM, memory network \\ \hline
\multirow{5}{0.5in}{Malware Detection}       & \cite{grosse2017adversarial}      & JSMA          & False negative    & White-box         & Targeted          & Individual    & Optimized         & Iterative         & $\ell_2$      & DREBIN        & 2-layer FC \\ \cline{2-12}
& \cite{anderson2017evading}    & Reinforcement Learning    & False negative    & Black-box     & Targeted          & Individual        & N/A           & Iterative         & N/A       & N/A                  & Gradient Boosted Decision Tree \\ \cline{2-12}
& \cite{hu2017generating}       & GAN       & False negative    & Black-box         & Targeted          & Individual    & N/A         & Iterative         & N/A      & malwr        & Multi-layer Perceptron \\ \cline{2-12}
& \cite{anderson2016deepdga}    & GAN       & False negative    & Black-box         & Targeted          & Individual    & N/A         & Iterative         & N/A      & Alexa Top 1M        & Random Forest \\ \cline{2-12}
& \cite{xu2016automatically}    & Generic Programming       & False negative    & Black-box         & Targeted          & Individual    & N/A         & Iterative         & N/A      & Contagio       & Random Forest, SVM \\ \hline
\end{tabularx}
\end{minipage}
}
\end{table*}

\subsection{Reinforcement Learning}
Deep neural networks have been used in reinforcement learning by training policies on raw input (\eg, images). \cite{huang2017adversarial, kos2017delving} generated adversarial examples on deep reinforcement learning policies. Since the inherent intensive computation of reinforcement learning, both of them performed fast \textit{One-time attack}.

Huang \etal applied \textit{FGSM} to attack deep reinforcement learning networks and algorithms~\cite{huang2017adversarial}: deep Q network (DQN), trust region policy optimization(TRPO), and asynchronous advantage actor-critic (A3C)~\cite{mnih2016asynchronous}. Similarly to~\cite{goodfellow2014explaining}, they added small perturbations on the input of policy by calculating the gradient of the cross-entropy loss function: $\nabla_x J(\theta, x, \ell))$. Since DQN does not have stochastic policy input, softmax of Q-values is considered to calculate the loss function.
They evaluated adversarial examples on four Atari 2600 games with three norm constraints $\ell_1, \ell_2, \ell_{\infty}$. They found \textit{Huang's Attack} with $\ell_1$ norm conducted a successful attack on both \textit{White-box attack} and \textit{Black-box attack} (no access to the training algorithms, parameters, and hyper-parameters).

\cite{kos2017delving} used \textit{FGSM} to attack A3C algorithm and Atari Pong task. \cite{kos2017delving} found that injecting perturbations in a fraction of frames is sufficient.

\subsection{Generative Modeling}
\begin{figure}[!t]
\centering
\includegraphics[width=0.85\linewidth]{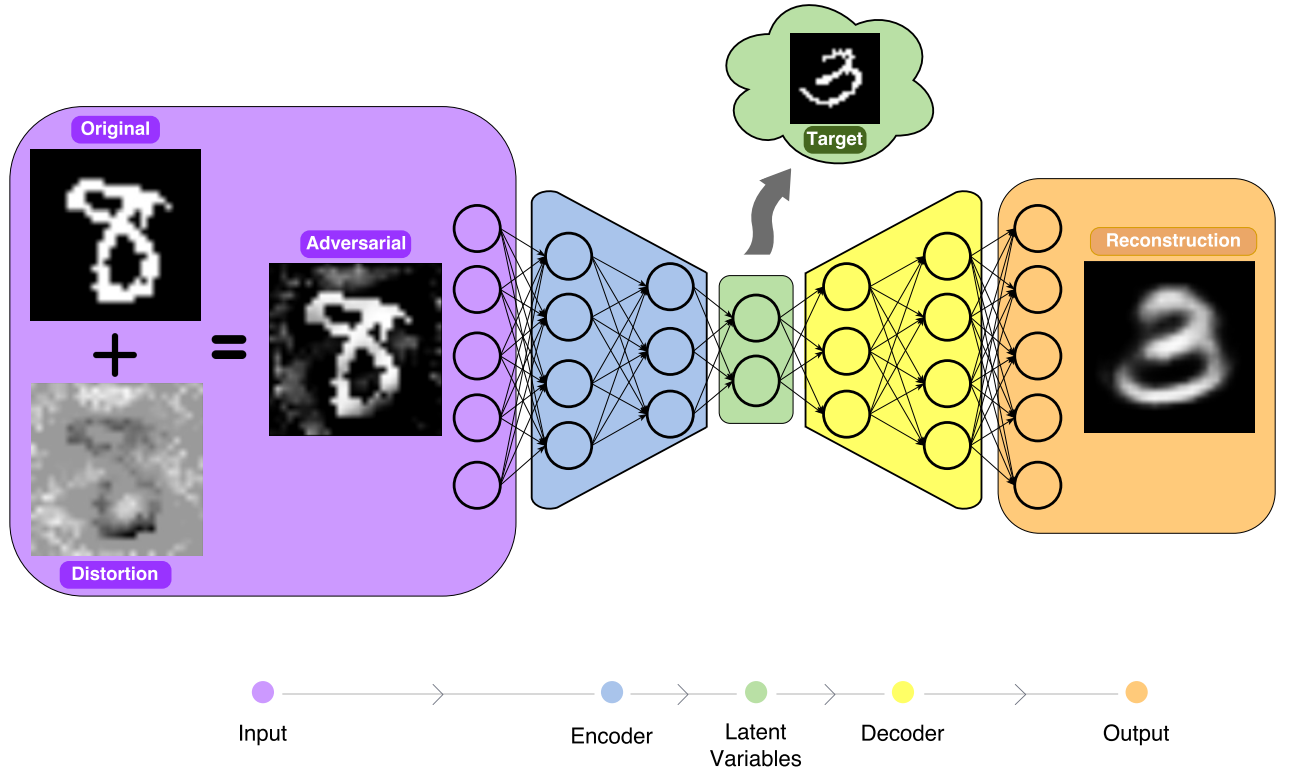}
\caption{Adversarial attacks for autoencoders~\cite{tabacof2016adversarial}. Perturbations are added to the input the encoder. After encoding and decoding, the decoder will output an adversarial image presenting an incorrect class}
\label{fig:autoencoder}
\vspace{-15pt}
\end{figure}

Kos \etal~\cite{kos2017adversarial} and Tabacof \etal~\cite{tabacof2016adversarial} proposed adversarial examples for generative models. An adversary for autoencoder can inject perturbations into the input of encoder and generate a targeted class after decoding.  Figure~\ref{fig:autoencoder} depicts a targeted adversarial example for an autoencoder. Adding perturbations on the input image of the encoder can misguide the autoencoder by making decoder to generating a targeted adversarial output image.

Kos \etal described a scenario to apply adversarial examples against autoencoder. Autoencoders can be used to compress data by an encoder and decompress by a decoder. For example, Toderici~\etal use RNN-based AutoEncoder to compress image~\cite{toderici2016full}. Ledig~\etal used GAN to super-resolve images~\cite{ledig2016photo}. Adversaries can leverage autoencoder to reconstruct an adversarial image by adding perturbation to the input of the encoder.

Tabacof \etal used \textit{Feature Adversary} Attack against AE and VAE. The adversarial examples were formulated as follows~\cite{tabacof2016adversarial}:
\begin{equation}
\begin{aligned}
& \min_{\eta} && D(z_{x'}, z_x) + c \|\eta\| \\
& s.t. && x' \in [L, U] \\
&&& z_{x'} = Encoder(x') \\ 
&&& z_x = Encoder(x) ,
\end{aligned}
\end{equation}
where $D(\cdot)$ is the distance between latent encoding representation $z_x$ and $z_{x'}$. Tabacof \etal chose KL-divergence to measure $D(\cdot)$ in~\cite{tabacof2016adversarial}. They tested their attacks on the MNIST and SVHN dataset and found that generating adversarial examples for autoencoder is much harder than for classifiers. VAE is even slightly more robust than deterministic autoencoder. 

Kos \etal extended Tabacof \etal's work by designing another two kinds of distances. Hence, the adversarial examples can be generated by optimizing: 
\begin{equation}
\min_{\eta} \quad  c\|\eta\|+J(x', l').
\end{equation}
The loss function $J$ can be cross-entropy (refer to ``Classifier Attack'' in~\cite{kos2017adversarial}), VAE loss function (``$L_{VAE}$ Attck''), and distance between the original latent vector $z$ and modified encoded vector $x'$ (``Latent Attack'', similar to Tabacof \etal's work~\cite{tabacof2016adversarial}). They tested VAE and VAE-GAN~\cite{larsen2015autoencoding} on the MNIST, SVHN, and CelebA datasets. In their experimental results, ``Latent Attack'' achieved the best result.

\subsection{Face Recognition}
Deep neural network based Face Recognition System (FRS) and Face Detection System have been widely deployed in commercial products due to their high performance. \cite{sharif2016accessorize} first provided a design of eyeglass frames to attack a deep neural network based FRS~\cite{parkhi2015deep}, which composes 11 blocks with 38 layers and one \textit{triplet loss} function for feature embedding. Based on the \textit{triplet loss} function, \cite{sharif2016accessorize} designed a \textit{softmaxloss} function:
\begin{equation}
J(x) = -\log \bigg(\frac{e^{<h_{c_x}, f(x)>}}{\sum_{c=1}^m e^{<h_c, f(x)>}}\bigg),
\end{equation}
where $h_c$ is a one-hot vector of class $c$, $<\cdot, \cdot>$ denotes inner product. Then they used \textit{L-BFGS Attack} to generate adversarial examples. 

\begin{figure}[!t]
    \centering
    \begin{subfigure}[b]{.3\columnwidth}
        \centering
        \includegraphics[width=0.75\columnwidth]{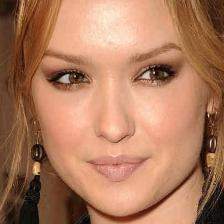}
    \end{subfigure}
    \hfill
    \begin{subfigure}[b]{.3\columnwidth}
        \centering
        \includegraphics[width=0.75\columnwidth]{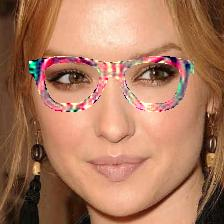}
    \end{subfigure}
    \hfill
    \begin{subfigure}[b]{.3\columnwidth}
        \centering
        \includegraphics[width=0.75\columnwidth]{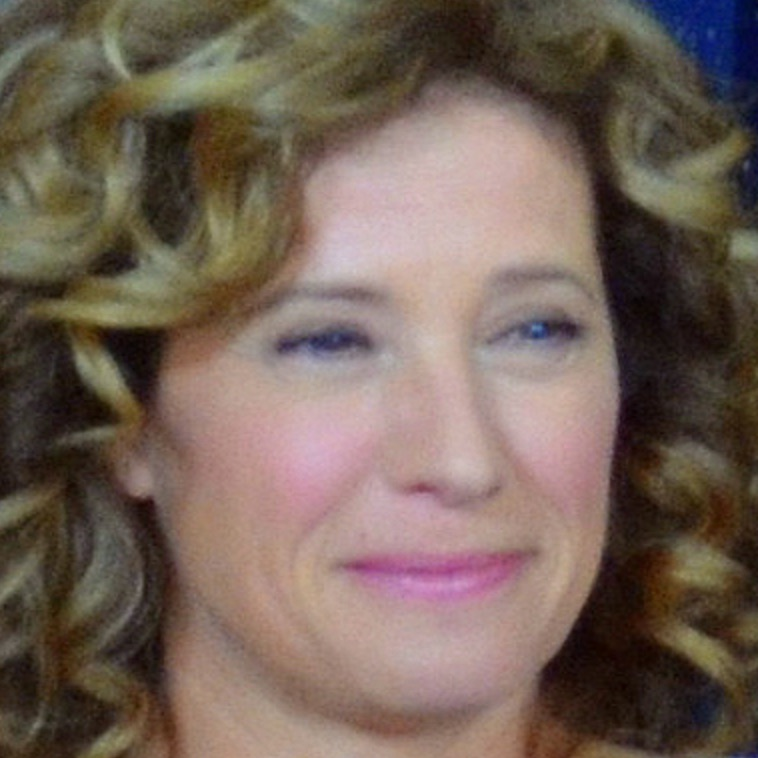}
    \end{subfigure}
    \caption{An example of adversarial eyeglass frame against Face Recognition System~\cite{sharif2016accessorize}}
    \label{fig:face}
    \vspace{-15pt}
\end{figure}

In a further step, \cite{sharif2016accessorize} implemented adversarial eyeglass frames to achieve attack in the physical world: the perturbations can only be injected into the area of eyeglass frames. They also enhanced the printability of adversarial images on the frame by adding a penalty of non-printability score (NPS) to the optimized objective. Similarly to \textit{Universal Perturbation}, they optimize the perturbation to be applied to a set of face images. They successfully dodged (non-targeted attack) against FRS (over 80 \% time) and  misguided FRS as a specific face (targeted attack) with a high success rate (depending on the target). Figure~\ref{fig:face} illustrates an example of adversarial eyeglass frames.

Leveraging the approach of printability, \cite{evtimov2017robust} proposed an attack algorithm, Robust Physical Perturbations ($RP_2$), to modify a stop sign as a speed limit sign)~\footnote{This method was shown not effective for standard detectors (YOLO and Faster RCNN) in~\cite{lu2017standard}.}. They changed the physical road signs by two kinds of attacks: 1) overlaying an adversarial road sign over a physical sign; 2) sticking perturbations on an existing sign. \cite{evtimov2017robust} included a non-printability score in the optimization objective to improve the printability.

\subsection{Object Detection}
\begin{figure}[!t]
\centering
\includegraphics[width=0.75\linewidth]{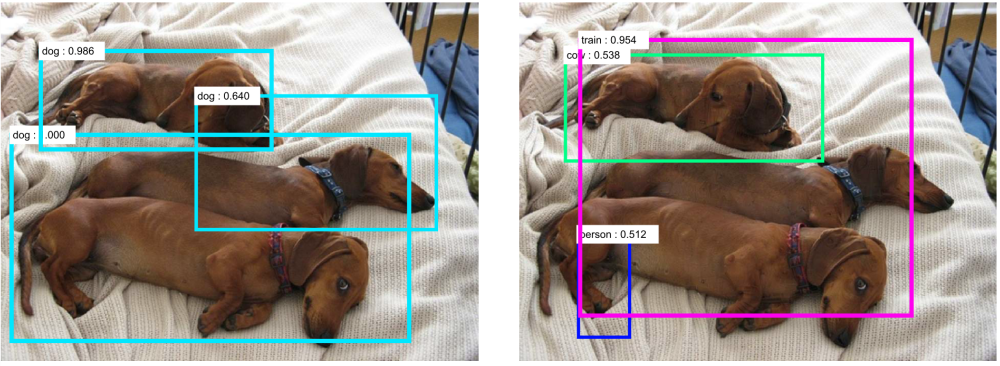}
\caption{An adversarial example for object detection task~\cite{xie2017adversarial}. Left: object detection on a clean image. Right: object detection on an adversarial image.} 
\label{fig:object}
\vspace{-15pt}
\end{figure}

The object detection task is to find the proposal of an object (bounding box), which can be viewed as an image classification task for every possible proposal. \cite{xie2017adversarial} proposed a universal algorithm called \textit{Dense Adversary Generation (DAG)} to generate adversarial examples for both object detection and semantic segmentation. The authors aimed at making the prediction (detection/segmentation) incorrect (non-targeted). Figure~\ref{fig:object} illustrates an adversarial example for the object detection task.

\cite{xie2017adversarial} defined $T={t_1, t_2, \ldots, t_N}$ as the recognition targets. For image classification, the classifier only needs one target -- entire image ($N=1$); For semantic segmentation, targets consist of all pixels ($N=\# of pixels$); For object detection, targets consist of all possible proposals ($N = (\# of pixels)^2$). Then the objective function sums up the loss from all targets. Instead of optimizing the loss from all targets, the authors performed an iterative optimization and only updated the loss for the targets correctly predicted in the previous iteration. The final perturbation sums up normalized perturbations in all iterations. To deal with a large number of targets for objective detection problem, the authors used regional proposal network (RPN)~\cite{ren2015faster} to generate possible targets, which greatly decreases the computation for targets in object detection. DAG also showed the capability of generating images which are unrecognizable to human but deep learning could predict (\textit{false positives}).

\subsection{Semantic Segmentation}
Image segmentation task can be viewed as an image classification task for every pixel. Since each perturbation is responsible for at least one pixel segmentation, this makes the space of perturbations for segmentation much smaller than that for image classification~\cite{hendrik2017universal}. \cite{xie2017adversarial,fischer2017adversarial,hendrik2017universal} generated adversarial examples against the semantic image segmentation task. However, their attacks are proposed under different scenarios. As we just discussed, \cite{xie2017adversarial} performed a non-targeted segmentation. \cite{fischer2017adversarial,hendrik2017universal} both performed a targeted segmentation and tried to removed a certain class by making deep learning model to misguide it as background classes.

\cite{fischer2017adversarial} generated adversarial examples by assigning pixels with the adversarial class that their nearest neighbor belongs to. The success rate was measured by the percentage of pixels of chosen class to be changed or of the rest classes to be preserved. 

\begin{figure}[!t]
\centering
\includegraphics[width=0.95\linewidth]{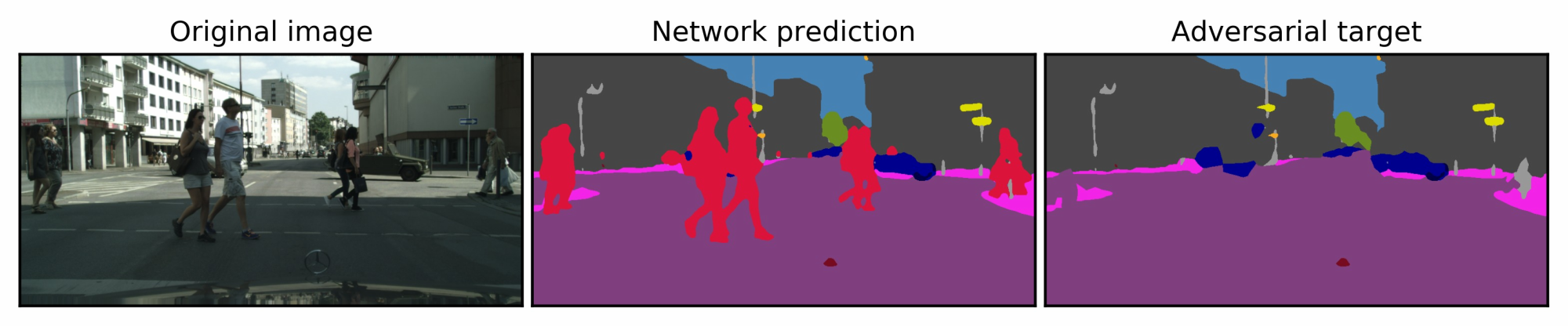}
\caption{Adversary examples of hiding pedestrians in the semantic segmentation task~\cite{hendrik2017universal}. Left image: original image; Middle image: the segmentation of the original image predicted by DNN; Right image: the segmentation of the adversarial image predicted by DNN.} 
\label{fig:segmentation}
\vspace{-15pt}
\end{figure}

\cite{hendrik2017universal} presented a method to generate universal adversarial perturbations against semantic image segmentation task. They assigned the primary objective of adversarial examples and hid the objects (\eg, pedestrians) while keeping the rest segmentation unchanged. Metzen \etal defined background classes and targeted classes (not targeted adversarial classes). Targeted classes are classes to be removed. Similar to \cite{fischer2017adversarial},  pixels which belong to the targeted classes would be assigned to their nearest background classes:

\begin{align}
\begin{split}
l_{ij}^{target} & = l_{i'j'}^{pred} \quad \forall (i,j) \in I_{targeted}, \\
l_{ij}^{target} & = l_{ij}^{pred} \quad \forall (i,j) \in I_{background}, \\
(i', j') & = \underset{(i', j') \in I_{background} }{\arg\min} \|i'-i\|+\|j'-j\|,\\
\end{split}
\end{align}
where $I_{targeted} = {(i,j)|f(x_{ij}) = l^*}$ denotes the area to be removed. Figure~\ref{fig:segmentation} illustrates an adversarial example to hide pedestrians.
They used \textit{ILLC} attack to solve this problem and also extended \textit{Universal Perturbation} method to get the universal perturbation. Their results showed the existence of universal perturbation for semantic segmentation task.

\subsection{Natural Language Processing (NLP)}
Many tasks in natural language processing can be attacked by adversarial examples. People usually generate adversarial examples by adding/deleting words in the sentences. 

The task of reading comprehension (a.k.a. question answering) is to read paragraphs and answer questions about the paragraphs. To generate adversarial examples that are consistent with the correct answer and do not confuse human, Jia and Liang added distracting (adversarial) sentences to the end of paragraph~\cite{jia2017adversarial}. They found that models for the reading comprehension task are overstable instead of oversensitivity, which means deep learning models cannot tell the subtle but critical difference in the paragraphs.

They proposed two kinds of methods to generate adversarial examples: 1) adding grammatical sentences similar to the question but not contradictory to the correct answer (AddSent); 2) adding a sentence with arbitrary English words (AddAny). \cite{jia2017adversarial} successfully fooled all the models (sixteen models) they tested on Stanford Question Answering Dataset (SQuAD)~\cite{rajpurkar2016squad}. The adversarial examples also have the capability of transferability and cannot be improved by adversarial training. However, the adversarial sentences require manpower to fix the errors in the sentences.

\cite{li2016understanding} aimed to fool a deep learning-based sentiment classifier by removing the minimum subset of words in the given text. Reinforcement learning was used to find an approximate subset, where the reward function was proposed as $\frac{1}{\|D\|}$ when the sentiment label changes, and $0$ otherwise. $\|D\|$ denotes the number of removing word set $D$. The reward function also included a regularizer to make sentence contiguous. 

The changes in\cite{jia2017adversarial,li2016understanding} can easily be recognized by humans. More natural adversarial examples for texture data was proposed by \textit{Natural GAN}~\cite{zhao2017generating} (Section~\ref{sec:natural_gan}).


\subsection{Malware Detection}
Deep learning has been used in static and behavioral-based malware detection due to its capability of detecting zero-day malware~\cite{yuan2014droid,dahl2013large,saxe2015deep,sun2017learning}. Recent studies generated adversarial malware samples to evade deep learning-based malware detection~\cite{grosse2017adversarial,anderson2017evading,hu2017generating,xu2016automatically}. 

\cite{grosse2017adversarial} adapted \textit{JSMA} method to attack Android malware detection model. \cite{xu2016automatically} evaded two PDF malware classifier, PDFrate and Hidost, by modifying PDF. \cite{xu2016automatically} parsed the PDF file and changed its object structure using genetic programming. The adversarial PDF file was then packed with new objects.

\cite{anderson2016deepdga} used GAN to generate adversarial domain names to evade detection of domain generation algorithms. \cite{hu2017generating} proposed a GAN based algorithm, MalGan, to generate malware examples and evade black-box detection. \cite{hu2017generating} used a substitute detector to simulate the real detector and leveraged the transferability of adversarial examples to attack the real detector. MalGan was evaluated by 180K programs with API features. However, \cite{hu2017generating} required the knowledge of features used in the model. \cite{anderson2017evading} used a large number of features (2,350) to cover the required feature space of portable executable (PE) files. The features included PE header metadata, section metadata, import \& export table metadata. \cite{anderson2017evading} also defined several modifications to generate malware evading deep learning detection. The solution was trained by reinforcement learning, where the evasion rate is considered as a reward.

\section{Countermeasures for Adversarial Examples}
\label{sec:counter}

Countermeasures for adversarial examples have two types of defense strategies: 1) \textit{reactive}: detect adversarial examples after deep neural networks are built; 2) \textit{proactive}: make deep neural networks more robust before adversaries generate adversarial examples. In this section, we discuss three reactive countermeasures (\textit{Adversarial Detecting}, \textit{Input Reconstruction}, and \textit{Network Verification}) and three proactive countermeasures (\textit{Network Distillation}, \textit{Adversarial (Re)training}, and \textit{Classifier Robustifying}). We will also discuss an ensembling method to prevent adversarial examples. Table~\ref{tab:counter} summarizes the countermeasures.

\begin{table}[!t]
\centering
\caption{Summary of Countermeasures for Adversarial Examples}
\label{tab:counter}
\begin{tabularx}{\linewidth}{|c|Y|Y|}
\hline
          & Defensive Strategies     & Representative Studies                                                                                                                                                                   \\ \hline
\multirow{3}{*}{Reactive}  & Adversarial Detecting    & \cite{bhagoji2017dimensionality,feinman2017detecting,gong2017adversarial,grosse2017statistical,metzen2017detecting,hendrycks2017early,lu2017safetynet,meng2017magnet,pang2017towards,lin2017detecting} \\ \cline{2-3}
          & Input Reconstruction     & \cite{gu2014towards,meng2017magnet,song2017pixeldefend}                                                                                                                                \\ \cline{2-3}
          & Network Verification     & \cite{katz2017reluplex,gopinath2017deepsafe,katz2017towards}                                                                                                                                                                \\ \hline
\multirow{3}{*}{Proactive} & Network Distillation     & \cite{papernot2016distillation}                                                                                                                                                        \\ \cline{2-3}
          & \multirow{2}{*}{Adversarial (Re)Training} & \cite{goodfellow2014explaining,huang2015learning,kurakin2017scale,tramer2017ensemble,dong2017towards,wu2017adversarial}                                                                \\ \cline{2-3}
          & Classifier Robustifying  & \cite{bradshaw2017adversarial,abbasi2017robustness}                                                                                                                                    \\ \hline
\end{tabularx}
\end{table}


\subsection{Network Distillation}
\label{sec:distillation}
\begin{figure}[!t]
\centering
\includegraphics[width=0.8\linewidth]{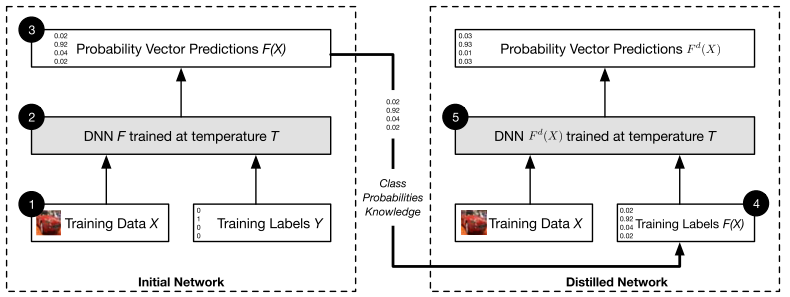}
\caption{Network distillation of deep neural networks~\cite{papernot2016distillation}} 
\label{fig:distillation}
\vspace{-15pt}
\end{figure}

Papernot \etal used network distillation to defend deep neural networks against adversarial examples~\cite{papernot2016distillation}. Network distillation was originally designed to reduce the size of deep neural networks by transferring knowledge from a large network to a small one~\cite{ba2014deep,hinton2015distilling} (Figure~\ref{fig:distillation}). The probability of classes produced by the first DNN is used as inputs to train the second DNN. The probability of classes extracts the knowledge learned from the first DNN. Softmax is usually used to normalize the last layer of DNN and produce the probability of classes. The softmax output of the first DNN, also the input of the next DNN, can be described as:
\begin{equation}
q_i = \frac{exp(z_i/T)}{\sum_j exp(z_j/T)},
\end{equation}
where $T$ is a \textit{temperature}  parameter to control the level of knowledge distillation. In deep neural networks, \textit{temperature} $T$ is set to $1$. When $T$ is large, the output of softmax will be vague (when $T \rightarrow \infty$, the probability of all classes $\rightarrow \frac{1}{m}$). When $T$ is small, only one class is close to $1$ while the rest goes to $0$. This schema of network distillation can be duplicated several times and connects several deep neural networks.

In~\cite{papernot2016distillation}, network distillation extracted knowledge from deep neural networks to improve robustness. The authors found that attacks primarily aimed at the sensitivity of networks and then proved that using high-temperature softmax reduced the model sensitivity to small perturbations. \textit{Network Distillation} defense was tested on the MNIST and CIFAR-10 datasets and reduced the success rate of \textit{JSMA} attack by 0.5\% and 5\% respectively. ``Network Distillation'' also improved the generalization of the neural networks.

\subsection{Adversarial (Re)training}
Training with adversarial examples is one of the countermeasures to make neural networks more robust. 
Goodfellow \etal~\cite{goodfellow2014explaining} and Huang \etal~\cite{huang2015learning} included adversarial examples in the training stage. They generated adversarial examples in every step of training and inject them into the training set. \cite{goodfellow2014explaining,huang2015learning} showed that adversarial training improved the robustness of deep neural networks. Adversarial training could provide regularization for deep neural networks~\cite{goodfellow2014explaining} and improve the precision as well~\cite{wu2017adversarial}. 

\cite{goodfellow2014explaining} and \cite{huang2015learning} were evaluated only on the MNIST dataset. A comprehensive analysis of adversarial training methods on the ImageNet dataset was presented in~\cite{kurakin2017scale}. They used half adversarial examples and half origin examples in each step of training. From the results, \textit{adversarial training} increased the robustness of neural networks for \textit{one-step attacks} (\eg, \textit{FGSM}) but would not help under \textit{iterative attacks} (\eg, \textit{BIM} and \textit{ILLC} methods). \cite{kurakin2017scale} suggested that adversarial training is used for regularization only to avoid overfitting (\eg, the case in \cite{goodfellow2014explaining} with the small MNIST dataset).

\cite{tramer2017ensemble} found that the adversarial trained models on the MNIST and ImageNet datasets are more robust to white-box adversarial examples than to the transferred examples (black-box). 

\cite{dong2017towards} minimized both the cross-entropy loss and internal representation distance during adversarial training, which can be seen as a defense version of \textit{Feature Adversary}.

To deal with the transferred black-box model, \cite{tramer2017ensemble} proposed \textit{Ensembling Adversarial Training} method that trained the model with adversarial examples generated from multiple sources: the models being trained and also pre-trained external models.


\subsection{Adversarial Detecting}
Many research projects tried to detect adversarial examples in the testing stage~\cite{bhagoji2017dimensionality,feinman2017detecting,gong2017adversarial,grosse2017statistical,metzen2017detecting,hendrycks2017early,lu2017safetynet,meng2017magnet,lin2017detecting}.

\cite{gong2017adversarial,metzen2017detecting,lu2017safetynet} trained deep neural network-based binary classifiers as detectors to classify the input data as a legitimate (clean) input or an adversarial example.
Metzen \etal created a detector for adversarial examples as an auxiliary network of the original neural network~\cite{metzen2017detecting}. The detector is a small and straightforward neural network predicting on binary classification, \ie, the probability of the input being adversarial. 
SafetyNet~\cite{lu2017safetynet} extracted the binary threshold of each ReLU layer's output as the features of the adversarial detector and detects adversarial images by an RBF-SVM classifier. The authors claimed that their method is hard to be defeated by adversaries even when adversaries know the detector, since it is difficult for adversaries to find an optimal value, for both adversarial examples and new features of SafetyNet detector.
\cite{grosse2017statistical} added an outlier class to the original deep learning model. The model detected the adversarial examples by classifying it as an outlier. They found that the measurement of maximum mean discrepancy (MMD) and energy distance (ED) could distinguish the distribution of adversarial datasets and clean datasets.

\cite{feinman2017detecting} provided a Bayesian view of detecting adversarial examples. \cite{feinman2017detecting} claimed that the uncertainty of adversarial examples is higher than the clean data. Hence, they deployed a Bayesian neural network to estimate the uncertainty of input data and distinguish adversarial examples and clean input data based on uncertainty estimation. 

Similarly, \cite{meng2017magnet} used probability divergence (Jensen-Shannon divergence) as one of its detectors. \cite{hendrycks2017early} showed that after whitening by Principal Component Analysis (PCA), adversarial examples have different coefficients in low-ranked components.

\cite{song2017pixeldefend} trained a PixelCNN neural network~\cite{salimans2017pixelcnn} and found that the distribution of adversarial examples is different from clean data. They calculated p-value based on the rank of PixelCNN and rejected adversarial examples using the p-values. The results showed that this approach could detect \textit{FGSM}, \textit{BIM}, \textit{DeepFool}, and \textit{C\&W} attack.

\cite{pang2017towards} trained neural networks with ``reverse cross-entropy'' to better distinguish adversarial examples from clean data in the latent layers and then detected adversarial examples using a method called ``Kernel density'' in the testing stage. The ``reverse cross-entropy'' made the deep neural network to predict with high confidence on the true class and uniform distribution on the other classes. In this way, the deep neural network was trained to map the clean input close to a low-dimensional manifold in the layer before softmax. This brought great convenience for further detection of adversarial examples.

\cite{lin2017detecting} leveraged multiple previous images to predict future input and detect adversarial examples, in the task of reinforcement learning. 


However, Carlini and Wagner summarized most of these adversarial detecting methods (\cite{bhagoji2017dimensionality,feinman2017detecting,gong2017adversarial,grosse2017statistical,metzen2017detecting,hendrycks2017early}) and showed that these methods could not defend against their previous attack \textit{C\&W's Attack} with slight changes of loss function~\cite{carlini2017adversarial,carlini2017magnet}.

\subsection{Input Reconstruction}
Adversarial examples can be transformed to clean data via reconstruction. After transformation, the adversarial examples will not affect the prediction of deep learning models. 
Gu and Rigazio proposed a variant of autoencoder network with a penalty, called deep contractive autoencoder, to increase the robustness of neural networks~\cite{gu2014towards}. A denoising autoencoder network is trained to encode adversarial examples to original ones to remove adversarial perturbations. 
\cite{meng2017magnet} reconstructed the adversarial examples by 1) adding Gaussian noise or 2) encoding them with autoencoder as a plan B in MagNet~\cite{meng2017magnet}(Section~\ref{sec:ens_defense}).

\textit{PixelDefend} reconstructed the adversarial images back to the training distribution~\cite{song2017pixeldefend} using PixelCNN. \textit{PixelDefend} changed all pixels along each channel to maximize the probability distribution:
\begin{equation}
\begin{aligned}
&\max_{x'} &&\mathcal{P}_t(x')\\
&s.t. && \|x'-x\|_{\infty} \leq \epsilon_{defend},
\end{aligned}
\end{equation}
where $\mathcal{P}_t$ denotes the training distribution, $\epsilon_{defend}$ controls the new changes on the adversarial examples. \textit{PixelDefend} also leveraged adversarial detecting, so that if an adversarial example is not detected as malicious, no change will be made to the adversarial examples ($\epsilon_{defend} = 0$).

\subsection{Classifier Robustifying}
\cite{bradshaw2017adversarial,abbasi2017robustness} design robust architectures of deep neural networks to prevent adversarial examples.

Due to the uncertainty from adversarial examples, Bradshaw \etal leveraged Bayesian classifiers to build more robust neural networks~\cite{bradshaw2017adversarial}. Gaussian processes (GPs) with RBF kernels were used to provide uncertainty estimation. The proposed neural networks were called \textit{Gaussian Process Hybrid Deep Neural Networks (GPDNNs)}. GPs expressed the latent variables as a Gaussian distribution parameterized by the functions of mean and covariance and encoded them with RBF kernels. \cite{bradshaw2017adversarial} showed that GPDNNs achieved comparable performance with general DNNs and more robust to adversarial examples. The authors claimed that GPDNNs ``know when they do not know.''


\cite{abbasi2017robustness} observed that adversarial examples usually went into a small subset of incorrect classes. \cite{abbasi2017robustness} separated the classes into sub-classes and ensembled the result from all sub-classes by voting to prevent adversarial examples misclassified. 

\subsection{Network Verification}
\label{sec:verify}
Verifying properties of deep neural networks is a promising solution to defend adversarial examples, because it may detect the new unseen attacks. \textit{Network verification} checks the properties of a neural network: whether an input violates or satisfies the property.


Katz \etal proposed a verification method for neural networks with ReLU activation function, called Reluplex~\cite{katz2017reluplex}. They used Satisfiability Modulo Theory (SMT) solver to verify the neural networks. The authors showed that within a small perturbation, there was no existing adversarial example to misclassify the neural networks. They also proved that the problem of network verification is NP-complete. Carlini \etal extended their assumption of ReLU function by presenting $\max(x,y)=ReLU(x-y)+y$ and $\|x\|=ReLU(2x)-x$~\cite{carlini2017ground}. However, Reluplex runs very slow due to the large computation of verifying the networks and only works for DNNs with several hundred nodes~\cite{katz2017towards}. \cite{katz2017towards} proposed two potential solutions: 1) prioritizing the order of checking nodes 2) sharing information of verification. 

Instead of checking each point individually, Gopinath \etal proposed \textit{DeepSafe} to provide safe regions of a deep neural network~\cite{gopinath2017deepsafe} using Reluplex. They also introduced \textit{targeted robustness} a safe region only regarding a targeted class.

\subsection{Ensembling Defenses}
\label{sec:ens_defense}
Due to the multi-facet of adversarial examples, multiple defense strategies can be performed together (parallel or sequential) to defend adversarial examples. 

Aforementioned \textit{PixelDefend}~\cite{song2017pixeldefend} is composed of an adversarial detector and an ``input reconstructor'' to establish a defense strategy.

\begin{figure}[!t]
\centering
\includegraphics[width=0.85\linewidth]{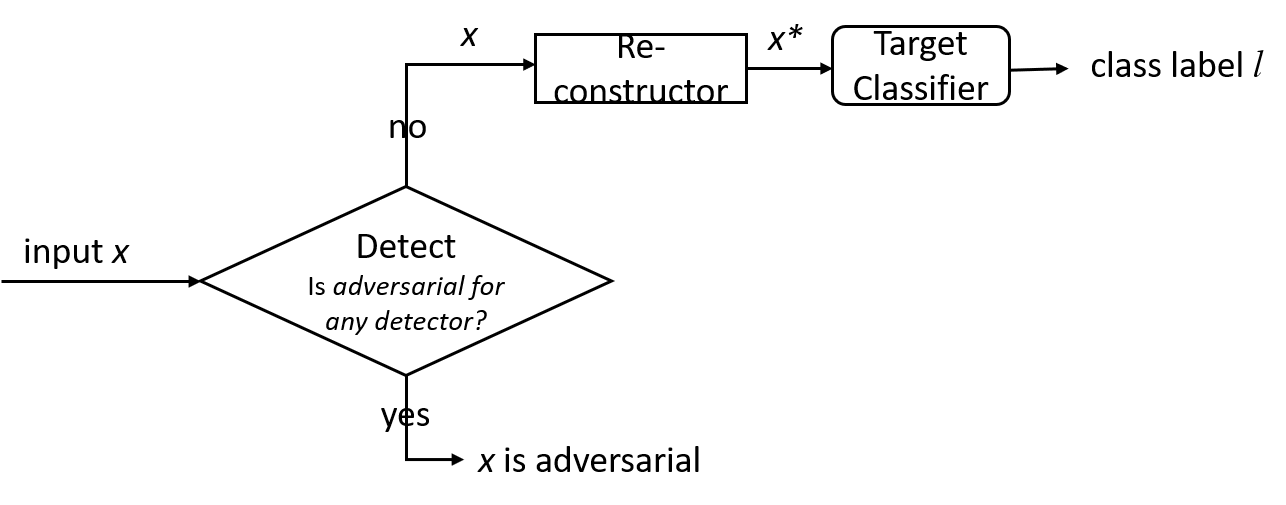}
\caption{MagNet workflow: one or more detectors first detects if input $x$ is adversarial; If not, reconstruct $x$ to $x^*$ before feeding it to the classifier. (modified from~\cite{meng2017magnet})} 
\label{fig:magnet}
\vspace{-15pt}
\end{figure}

MagNet included one or more detectors and a reconstructor (``reformer'' in the paper) as Plan A and Plan B~\cite{meng2017magnet}. The detectors are used to find the adversarial examples which are far from the boundary of the manifold. In~\cite{meng2017magnet}, they first measured the distance between input and encoded input and also the probability divergence (Jensen-Shannon divergence) between softmax output of input and encoded input. The adversarial examples were expected a large distance and probability divergence. To deal with the adversarial examples close to the boundary, MagNet used a reconstructor built by neural network based autoencoders. The reconstructor will map adversarial examples to legitimate examples. Figure~\ref{fig:magnet} illustrates the workflow of the defense of two phases.

After investigating several defensive approaches, \cite{he2017adversarial} showed that the ensemble of those defensive approaches does not make the neural networks strong.



\subsection{Summary}
Almost all defenses are shown to be effective only for part of attacks. They tend not to be defensive for some strong (fail to defend) and unseen attacks. Most defenses target adversarial examples in the computer vision task. However, with the development of adversarial examples in other areas, new defenses for these areas, especially for safety-critical systems, are urgently required. 

\section{Challenges and Discussions}
\label{sec:analysis}
In this section, we discuss the current challenges and the potential solutions for adversarial examples. Although many methods and theorems have been proposed and developed recently, a lot of fundamental questions need to be well explained and many challenges need to be addressed. The reason for the existence of adversarial examples is an interesting and one of the most fundamental problems for both adversaries and researchers, which exploits the vulnerability of neural networks and help defenders to resist adversarial examples. We will discuss the following questions in this section: Why do adversarial examples transfer? How to stop the transferability? Why are some defenses effective and others not? How to measure the strength of an attack as well as a defense? How to evaluate the robustness of a deep neural network against seen/unseen adversarial examples?

\subsection{Transferability}
\label{sec:transfer}
Transferability is a common property for adversarial examples. Szegedy \etal first found that adversarial examples generated against a neural network can fool the same neural networks trained by different datasets. Papernot \etal found that adversarial examples generated against a neural network can fool other neural networks with different architectures, even other classifiers trained by different machine learning algorithms~\cite{papernot2016transferability}. Transferability is critical for Black-Box attacks where the victim deep learning model and the training dataset are not accessible. Attackers can train a substitute neural network model and then generate adversarial examples against substitute model. Then the victim model will be vulnerable to these adversarial examples due to transferability. 
From a defender's view, if we hinder transferability of adversarial examples, we can defend all white-box attackers who need to access the model and require transferability. 

We define the transferability of adversarial examples in three levels from easy to hard: 1) transfer among the same neural network architecture trained with different data; 2) transfer among different neural network architectures trained for the same task; 3) transfer among deep neural networks for different tasks. To our best knowledge, there is no existing solution on the third level yet (for instance, transfer an adversarial image from object detection to semantic segmentation).


Many studies examined transferability to show the ability of adversarial examples~\cite{szegedy2013intriguing,goodfellow2014explaining}. Papernot \etal studied the transferability between conventional machine learning techniques (\ie, logistic regression, SVM, decision tree, kNN) and deep neural networks. They found that adversarial examples can be transferred between different parameters, training dataset of a machine learning models and even across different machine learning techniques. 

Liu \etal investigated transferability of targeted and non-targeted adversarial examples on complex models and large datasets (\eg, the ImageNet dataset)~\cite{liu2017delving}. They found that non-targeted adversarial examples are much more transferable than targeted ones. They observed that the decision boundaries of different models aligned well with each other. Thus they proposed \textit{Model-Based Ensembling Attack} to create transferable targeted adversarial examples.

Tram\`er \etal found that the distance to the model's decision boundary is on average larger than the distance between two models' boundaries in the same direction~\cite{tramer2017space}. This may explain the existence of transferability of adversarial examples. Tram\`er \etal also claimed that transferability might not be an inherent property of deep neural networks by showing a counter-example. 



\subsection{The existence of Adversarial Examples}
The reason for the existence of adversarial examples is still an open question. Are adversarial examples an inherent property of deep neural networks? Are adversarial examples the ``Achilles' heel'' of deep neural networks with high performance? Many hypotheses have been proposed to explain the existence.

\textbf{Data incompletion}
One assumption is that adversarial examples are of low probability and low test coverage of corner cases in the testing dataset~\cite{szegedy2013intriguing, pei2017deepxplore}. From training a PixelCNN, \cite{song2017pixeldefend} found that the distribution of adversarial examples was different from clean data. 
Even for a simple Gaussian model, a robust model can be more complicated and requires much more training data than that of a ``standard'' model~\cite{schmidt2018adversarially}.

\textbf{Model capability}
Adversarial examples are a phenomenon not only for deep neural networks but also for all classifiers~\cite{fawzi2015fundamental,papernot2016transferability}. \cite{goodfellow2014explaining} suggested that adversarial examples are the results of models being too linear in high dimensional manifolds. \cite{tanay2016boundary} showed that in the linear case, the adversarial examples exist when the decision boundary is close to the manifold of the training data. 

Contrary to \cite{goodfellow2014explaining}, \cite{fawzi2015fundamental} believed that adversarial examples are due to the ``low flexibility'' of the classifier for certain tasks.  Linearity is not an ``obvious explanation''~\cite{sabour2016adversarial}. \cite{tabacof2016exploring} blamed adversarial examples for the sparse and discontinuous manifold which makes classifier erratic. 

\textbf{No robust model} 
\cite{dong2017towards} suggested that the decision boundaries of deep neural networks are inherently incorrect, which do not detect semantic objects. 
\cite{fawzi2018adversarial} showed that if a dataset is generated by a smooth generative model with large latent space, there is no robust classifier to adversarial examples. Similarly, \cite{gilmer2018adversarial} prove that if a model is trained on a sphere dataset and misclassifies a small part of the dataset, then there exist adversarial examples with a small perturbation.





In addition to adversarial examples for image classification task, as discussed in Section~\ref{sec:app}, adversarial examples have been generated in various applications. Many of them deployed utterly different methods. Some applications can use the same method used in image classification task. However, some need to propose a novel method. Current studies on adversarial examples mainly focus on image classification task. No existing paper explains the relationship among different applications and existence of a universal attacking/defending method to be applied to all the applications.

\subsection{Robustness Evaluation}
\label{sec:analysis_evaluation}
The competition between attacks and defenses for adversarial examples becomes an ``arms race'': a defensive method that was proposed to prevent existing attacks was later shown to be vulnerable to some new attacks, and vice versa~\cite{grosse2017statistical,carlini2017ground}. Some defenses showed that they could defend a particular attack, but later failed with a slight change of the attack~\cite{feinman2017detecting,bhagoji2017dimensionality}. Hence, the evaluation on the robustness of a deep neural network is necessary. For example, \cite{fawzi2015analysis} provided an upper bound of robustness for linear classifier and quadratic classifier. The following problems for robustness evaluation of deep neural networks require further exploration.

1) \textbf{A methodology for evaluation on the robustness of deep neural networks}: 
Many deep neural networks are planned to be deployed in safety-critical settings. Defending only existing attacks is not sufficient. Zero-day (new) attacks would be more harmful to deep neural networks. A methodology for evaluating the robustness of deep neural networks is required, especially for zero-day attacks, which helps people understand the confidence of model prediction and how much we can rely on them in the real world. 
\cite{bastani2016measuring,huang2017safety,katz2017reluplex,carlini2017ground} conducted initial studies on the evaluation. Moreover, this problem lies not only in the performance of deep neural network models but also in the confidentiality and privacy.  

2) \textbf{A benchmark platform for attacks and defenses}: 
Most attacks and defenses described their methods without publicly available code, not to mention the parameters used in their methods. This brings difficulties for other researchers to reproduce their solutions and provide the corresponding attacks/defenses. For example, Carlini tried his best to ``find the best possible defense parameters + random initialization''\footnote{Code repository used in~\cite{carlini2017adversarial}: \url{https://github.com/carlini/nn_breaking_detection}}. Some researchers even drew different conclusions because of different settings in their experiments. If there exists any benchmark, where both adversaries and defenders conduct experiments in a uniform way (i.e., the same threat model, dataset, classifier, attacking/defending approach), we can make a more precise comparison between different attacking and defending techniques.

Cleverhans~\cite{papernot2017cleverhans} and Foolbox~\cite{rauber2017foolbox} are open-source libraries to benchmark the vulnerability of deep neural networks against adversarial images. They build frameworks to evaluate the attacks. However, defensive strategies are missing in both tools. Providing a dataset of adversarial examples generated by different methods will make it easy for finding the blind point of deep neural networks and developing new defense strategies. This problem also occurs in other areas in deep learning. 

Google Brain organized three competitions in NIPS 2017 competition track, including targeted adversarial attack, non-targeted adversarial attack, and defense against adversarial attack~\cite{kurakin2018adversarial}. The dataset in the competition consisted of a set of images never used before and manually labeled the images, 1,000 images for development and 5,000 images for final testing. The submitted attacks and competitions are used as benchmarks to evaluate themselves. The adversarial attacks and defenses are scored by the number of runs to fool the defenses/correctly classify images.

\begin{figure}[!t]
\centering
\includegraphics[width=0.85\linewidth]{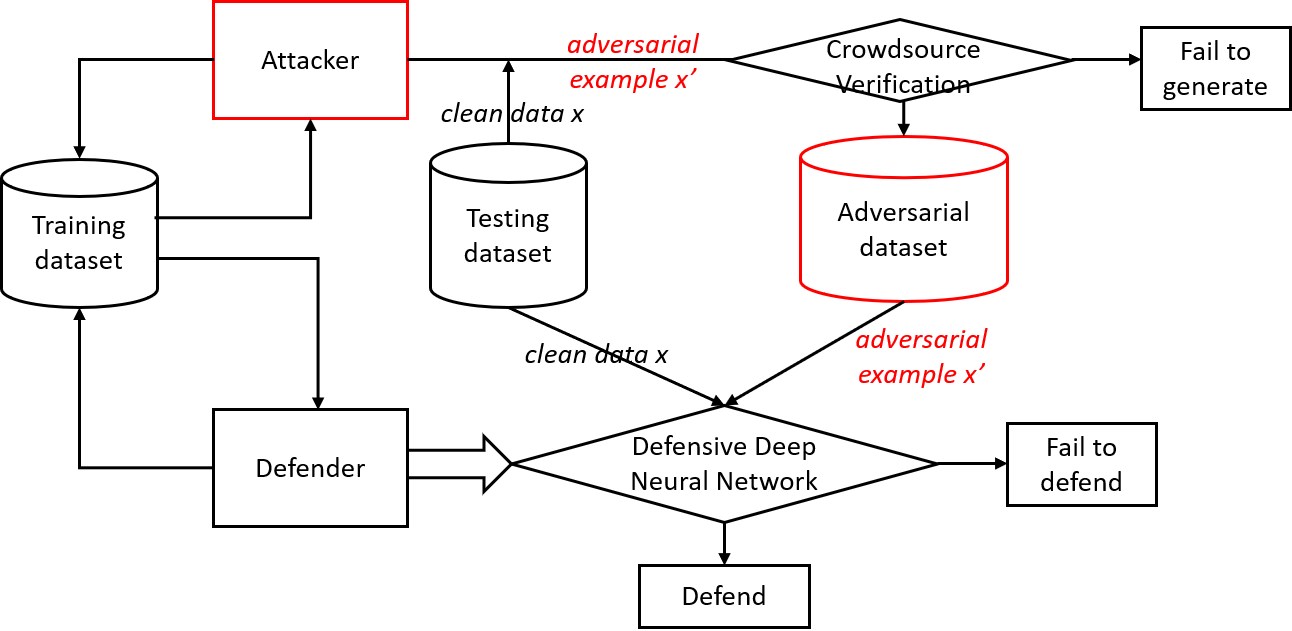}
\caption{Workflow of a benchmark platform for attackers and defenders: 1) attackers and defenders update/train their strategies on training dataset; 2) attackers generate adversarial examples on the clean data; 3) the adversarial examples are verified by crowdsourcing whether recognizable to human; 4) defenders generate a deep neural network as a defensive strategy; 5) evaluate the defensive strategy.}
\label{fig:benchmark}
\vspace{-15pt}
\end{figure}

We present workflow of a benchmark platform for attackers and defenders (Figure~\ref{fig:benchmark}).

3) \textbf{Various applications for robustness evaluation}: 
Similar to the existence of adversarial examples for various applications, a wide range of applications make it hard to evaluate the robustness, of a deep neural network architecture. How to compare methods generating adversarial example under different threat models? Do we have a universal methodology to evaluate the robustness under all scenarios? Tackling these unsolved problems is a future direction.


\section{Conclusion}
\label{sec:conclusion}
In this paper, we reviewed recent findings of adversarial examples in deep neural networks. We investigated existing methods for generating adversarial examples\footnote{Due to the rapid development of adversarial examples (attacks and defenses), we only considered the papers published before November 2017. We will update the survey with new methodologies and papers in our future work.}. A taxonomy of adversarial examples was proposed.  We also explored the applications and countermeasures for adversarial examples. 

This paper attempted to cover state-of-the-art studies for adversarial examples in the deep learning domain. Compared with recent work on adversarial examples, we analyzed and discussed current challenges and potential solutions in adversarial examples.




\bibliographystyle{IEEEtran}
\bibliography{bib/adversial,bib/deep,bib/privacy}


\end{document}